\documentclass[lettersize,journal]{IEEEtran}
\usepackage{amsmath,amsfonts}
\usepackage{algorithmic}
\usepackage{algorithm}
\usepackage{array}
\usepackage[caption=false,font=normalsize,labelfont=sf,textfont=sf]{subfig}
\usepackage{textcomp}
\usepackage{stfloats}
\usepackage{url}
\usepackage{verbatim}
\usepackage{graphicx}
\usepackage{cite}
\usepackage{comment}
\usepackage{natbib}

\usepackage{booktabs}

\def\BibTeX{{\rm B\kern-.05em{\sc i\kern-.025em b}\kern-.08em
    T\kern-.1667em\lower.7ex\hbox{E}\kern-.125emX}}
\usepackage{verbatim}

\usepackage[pagebackref=true,breaklinks=true,letterpaper=true,colorlinks,bookmarks=false]{hyperref}

\usepackage{lscape}
\usepackage{pdflscape}

\usepackage{longtable}
\usepackage{adjustbox}
\usepackage{geometry}
\usepackage{array}
\usepackage{graphicx} 
\usepackage{tcolorbox} 

\begin{document}

\title{From AI Technical Debt to Agentic Technical Debt: A Systematic Mapping of Root Causes and Manifestations in Agentic AI Systems}

\author{\IEEEauthorblockN{Muhammad Tukur$^a$$^,$$^d$$^*$, Hayatullahi Adeyemo$^b$, Tao Chen$^a$, Nour Ali$^c$,  Anis Zarrad$^a$, Marco Agus$^d$, Rick Kazman$^e$, Rami Bahsoon$^a$$^*$}

\IEEEauthorblockN{\textit{$^a$School of Computer Science, University of Birmingham, Edgbaston, UK} \\
\textit{$^b$Computing and Informatics, Bournemouth University, Bournemouth, UK}\\
\textit{$^c$Brunel University London, Uxbridge, UK}\\
\textit{$^d$College of Science and Engineering, HBKU, Qatar}\\
\textit{$^e$University of Hawaii, Honolulu, USA}\\
}}

\markboth{Journal of \LaTeX\ Class Files,~Vol.~14, No.~8, August~2026}%
{Shell \MakeLowercase{\textit{et al.}}: A Sample Article Using IEEEtran.cls for IEEE Journals}


\maketitle

\begin{abstract}
The emergence of Agentic AI systems, characterized by autonomous reasoning, multi-agent collaboration, tool orchestration, adaptive decision-making, and persistent memory, represents a fundamental shift from traditional AI pipelines to dynamic, self-directed software ecosystems. While AI Technical Debt (AITD) has been widely studied in conventional machine learning and software engineering contexts, existing models largely assume static, component-level architectures and fail to capture the dynamic, distributed, and emergent behaviors of agentic environments. This gap limits the systematic identification, analysis, and management of technical debt in modern autonomous AI systems.
This paper introduces the concept of \textit{Agentic Technical Debt (AgTD)}, defined as the forms of technical debt that emerge, accumulate, propagate, and amplify due to the autonomous, collaborative, and adaptive nature of Agentic AI systems. Building upon our prior systematic scoping review that identified 31 AITDs across seven root-cause categories, we employ a theory-informed transformation methodology to systematically reinterpret and map these debts to their corresponding manifestations in Agentic AI systems. Through direct transformation, contextual transformation, and manifestation expansion, we present the first systematic mapping of established AITDs to agentic manifestations, demonstrating how conventional debts evolve into dynamic system-level liabilities, including memory inconsistencies, orchestration fragility, cascading failures, and unsafe autonomous decision-making.
Our findings show that technical debt in Agentic AI extends beyond software artifacts to encompass agent behaviors, coordination mechanisms, adaptive reasoning, and interactions among agents, tools, and execution environments. We further analyze its implications for AI Trust, Risk, and Security Management (TRiSM), highlighting impacts on trustworthiness, governance, security, operational resilience, and the emerging challenge of Sustainability Technical Debt (SusTD). Overall, this work establishes AgTD as a foundational software engineering construct and provides a theory-informed transformation framework, a systematic taxonomy, and a research agenda for the quantification, monitoring, governance, mitigation, and sustainability-aware management of technical debt in autonomous multi-agent AI systems.
\end{abstract}


\section{Introduction}

Artificial Intelligence (AI) has undergone a rapid evolution from rule-based systems to data-driven machine learning and, more recently, to generative and large language model (LLM)-based systems~\cite{mundlamuri2025evolution}. Despite these advances, most traditional AI systems remain largely \textit{reactive}, operating within predefined pipelines and requiring significant human supervision~\cite{gawande2025reactive}. The emergence of \textit{Agentic AI} marks a fundamental shift in this paradigm, enabling systems to autonomously \textit{perceive, reason, act, and learn} in dynamic environments with minimal human intervention \cite{raheem2025agentic}. These systems are characterized by goal-directed behavior, multi-agent coordination, tool usage, and persistent memory, allowing them to solve complex, multi-step tasks across diverse domains such as healthcare~\cite{collaco2026role}, finance~\cite{chugh2025opportunities}, cybersecurity~\cite{kshetri2025transforming}, and industrial automation~\cite{pati2025agentic,bandi2025rise}.

Recent advances in agentic frameworks-leveraging technologies such as retrieval-augmented generation (RAG)~\cite{habib2025towards}, multi-agent orchestration~\cite{rafe2026orchestration}, and external tool integration-have significantly enhanced system capabilities. Agentic AI systems can now plan, execute, and adapt strategies in real time, often collaborating across multiple specialized agents to achieve shared objectives~\cite{hosseini2025role}. This shift from ``copilot'' assistance to ``autopilot'' autonomy has enabled substantial gains in productivity, efficiency, and scalability, while also introducing new challenges related to trust, safety, and governance \cite{hosseini2025role}. As these systems become increasingly embedded in critical infrastructures and decision-making processes, ensuring their reliability and robustness becomes paramount.

Parallel to these developments, the concept of \textit{Technical Debt (TD)} has been widely recognized as a critical factor affecting the long-term maintainability, reliability, and quality of software systems~\cite{cunningham1992wycash,ernst2021technical}. In the context of AI, \textit{AI Technical Debt (AITD)} extends this notion to encompass issues arising from data dependencies, model design, pipeline complexity, and operational processes~\cite{bogner2021characterizing, sculley2015hidden}. Prior research has identified a comprehensive set of AITDs and categorized them into root-cause classes such as data, model and code, architecture, operational processes, and testing. However, these studies largely assume \textit{static, pipeline-oriented AI systems}, where components are relatively isolated and system behavior is predictable.


The transition to Agentic AI fundamentally challenges these assumptions. While properties such as dynamicity, distribution, and interaction-driven behavior are also present in many distributed and self-adaptive software systems, Agentic AI introduces an additional layer of complexity through LLM-driven reasoning, autonomous decision-making, tool invocation, and agent-to-agent collaboration. These systems operate as socio-technical ecosystems of interacting agents that continuously perceive, reason, act, and adapt, creating evolving dependencies across models, tools, data sources, and execution environments~\cite{abou2025agentic}. As a result, technical debt in such systems no longer remains confined to individual components but instead \textit{emerges and propagates across the entire system}, often manifesting as coordination failures, cascading errors, memory inconsistencies, and unintended emergent behaviors. Moreover, the integration of external tools, APIs, and knowledge sources introduces additional attack surfaces, trust dependencies, and operational risks, further complicating the identification, management, and mitigation of technical debt in these environments~\cite{rashid2025securing}. Consequently, technical debt can arise not only from conventional software artifacts but also from agent behaviors, coordination mechanisms, and emergent interactions that are difficult to anticipate, govern, and control.

Despite the growing importance of Agentic AI, there is currently a lack of systematic understanding of how traditional AI technical debts evolve in these systems. Existing AITD taxonomies do not account for key characteristics of agentic systems, such as autonomy, multi-agent interaction, and persistent memory. Consequently, there is a critical need to revisit and extend the concept of technical debt to reflect the realities of modern AI systems.

To address this gap, this paper introduces the concept of \textit{Agentic Technical Debt (AgTD)}, defined as the forms of technical debt that arise, accumulate, and propagate due to the autonomous, collaborative, and adaptive nature of agentic AI systems. Building on our prior systematic review that identified 31 AITDs across seven root-cause categories~\cite{tukur2026aisafetysecuritytechnical}, we present a systematic mapping of these debts to their corresponding manifestations in agentic contexts. This mapping reveals how traditional debts-such as data debt~\cite{akgul2025aligning}, glue code~\cite{alahdab2019empirical}, hidden feedback loops~\cite{khritankov2021hidden}, and boundary erosion~\cite{chaudhary2018review}-transform into new system-level challenges, including memory corruption, orchestration fragility, cascading failures, and unsafe autonomous decision-making.

Beyond extending existing AITD concepts to agentic environments, this work also identifies \emph{Sustainability Technical Debt (SusTD)} as an emerging and underexplored dimension of AgTD. Owing to continuous reasoning, autonomous planning, persistent memory management, repeated inference, tool orchestration, and multi-agent collaboration, Agentic AI systems can incur substantial computational and energy demands~\cite{lee2026toward}. Consequently, architectural inefficiencies and suboptimal agent workflows may accumulate as sustainability-related debt, affecting energy efficiency, operational costs, and environmental impact throughout the system lifecycle. This observation broadens the scope of AgTD beyond conventional software quality concerns toward environmentally sustainable autonomous AI engineering.



The contributions of this paper are threefold:

\begin{itemize}
\item \textbf{Conceptual Contribution:} We introduce and formally define \textit{Agentic Technical Debt (AgTD)}, extending the traditional theory of technical debt to agentic AI systems. We characterize how autonomy, reasoning, tool usage, memory mechanisms, and multi-agent interactions create new debt accumulation pathways that are not adequately captured by existing AI technical debt frameworks.

\item \textbf{Empirical Contribution:} We present the first systematic mapping of 31 AITDs to their manifestations in agentic AI systems. Using a root-cause-based taxonomy, we identify how existing debt types evolve, propagate, and amplify within autonomous and collaborative agent ecosystems, revealing emerging debt patterns specific to agentic architectures.

\item \textbf{Analytical Contribution:} We analyze the implications of AgTD for Trust, Risk, and Security Management (TRiSM), examining its impact on explainability, governance, reliability, security, and operational resilience. We further highlight key research challenges and opportunities for monitoring, assessing, and mitigating technical debt in agentic AI environments.

\end{itemize}

By bridging the gap between traditional technical debt theory and emerging agentic AI systems, this work establishes a foundation for future research on the identification, quantification, governance, and mitigation of technical debt in autonomous, adaptive, and multi-agent environments.

The remainder of this paper is structured as follows. 
Section~\ref{sec:background} presents the background of the study. 
Section~\ref{sec:relatedwork} reviews the related work. 
Section~\ref{sec:methodology} describes the research methodology. 
Section~\ref{sec:agtd} introduces the formal definition of AgTDs and presents their systematic mapping and analysis. 
Section~\ref{sec:agtd_aitrism} discusses the implications of AgTDs for AI TRiSM. 
Finally, Section~\ref{sec:conclusion} concludes the paper and outlines future research directions.

\section{Background}
\label{sec:background}
This section provides the conceptual foundation for the study by contextualizing the evolution of Artificial Intelligence (AI) paradigms and their implications for software engineering practices. We begin by outlining the progression from traditional AI~\cite{aggarwal2025traditional} to Generative AI~\cite{fui2023generative} and, more recently, to Agentic AI~\cite{huang2025agentic}, highlighting their defining characteristics, capabilities, and key differences. Building on this, we revisit the concept of Technical Debt (TD) and examine how it has evolved from traditional software systems to AI-based systems TD, and further to Agentic AI systems TD. This progression establishes the theoretical basis for understanding how autonomy, interaction, and system-level dynamics fundamentally transform the nature, manifestation, and impact of technical debt in modern AI ecosystems.

\subsection{Artificial Intelligence, Generative AI, and Agentic AI}

Artificial Intelligence (AI) refers to computational systems capable of performing tasks that typically require human intelligence, including perception, reasoning, learning, and decision-making~\cite{jaboob2024artificial}. Traditional AI systems are primarily designed to execute predefined tasks through rule-based reasoning or machine learning models trained for specific objectives, such as classification, prediction, recommendation, or anomaly detection~\cite{davis1984origin,pasrija2022machine}. While these systems have achieved significant success across numerous domains, they generally operate within well-defined boundaries and rely heavily on human oversight for task specification and execution.

Generative AI (GenAI) extends traditional AI by enabling systems to generate novel content, including text, images, code, audio, and video~\cite{kumar2025fundamentals,parasuraman2024introduction}. Powered largely by foundation models and Large Language Models (LLMs), Generative AI has transformed human-computer interaction through natural language interfaces and advanced content creation capabilities~\cite{kaviyaraj2024generative}. Despite these advances, most generative systems remain fundamentally reactive, producing outputs in response to user prompts without independently pursuing objectives, coordinating actions, or managing long-term tasks~\cite{konstantinou2024leveraging}.

Agentic AI represents the next stage in the evolution of AI systems by moving beyond content generation toward autonomous goal execution~\cite{abou2025agentic}. Rather than functioning solely as predictive or generative models, agentic systems operate as intelligent entities capable of perceiving their environment, reasoning about objectives, planning actions, executing tasks, and adapting their behavior over time~\cite{acharya2025agentic,garg2025designing}. This shift transforms AI from a passive assistant into an active participant capable of independently pursuing complex goals with limited human intervention.

A defining characteristic of Agentic AI is \textit{autonomy}. Agentic systems can make decisions and initiate actions without requiring continuous user prompts. Given a high-level objective, an agent can decompose the problem into smaller tasks, determine an execution strategy, monitor progress, and adjust its behavior in response to changing conditions~\cite{widad2025evolution}. This capability enables the execution of long-horizon tasks that would otherwise require substantial human coordination.

Another key characteristic is \textit{goal-directed behavior}. Unlike traditional and generative systems that primarily respond to immediate inputs, agentic systems maintain explicit objectives and continuously evaluate their actions against desired outcomes. This allows them to proactively identify missing information, revise strategies, and pursue alternative solutions when obstacles arise~\cite{abou2025agentic}.

Agentic systems also rely heavily on \textit{reasoning and planning}. Modern agents leverage LLMs and other AI models to analyze context, evaluate alternatives, generate action plans, and perform multi-step decision-making processes~\cite{garg2025designing}. Planning mechanisms enable agents to sequence actions, allocate resources, and coordinate dependencies among tasks, significantly enhancing their ability to solve complex real-world problems.

A further distinguishing feature is \textit{tool orchestration}. Agentic AI systems are not limited to the knowledge encoded within their underlying models. Instead, they can dynamically invoke external tools, APIs, databases, search engines, software services, and domain-specific applications to acquire information or perform actions~\cite{chugh2025opportunities}. Through tool integration, agents can extend their capabilities beyond content generation and interact directly with operational environments.

Effective operation in complex environments also requires \textit{memory}. Unlike conventional LLM-based systems that rely primarily on limited context windows, agentic systems increasingly incorporate persistent memory mechanisms that enable them to retain knowledge, maintain state across interactions, and learn from previous experiences~\cite{collaco2026role}. Memory supports continuity, personalization, contextual awareness, and long-term task execution, allowing agents to make more informed decisions over time.

Finally, many agentic architectures employ \textit{multi-agent collaboration}. Instead of relying on a single monolithic agent, complex tasks can be distributed among multiple specialized agents that communicate, coordinate, and cooperate toward shared objectives~\cite{RAZA202671}. Such architectures improve scalability, modularity, and problem-solving effectiveness by enabling agents to assume specialized roles such as planning, execution, monitoring, verification, or domain-specific reasoning.

\begin{figure*}[t]
    \centering
    \includegraphics[width=1.0\textwidth]{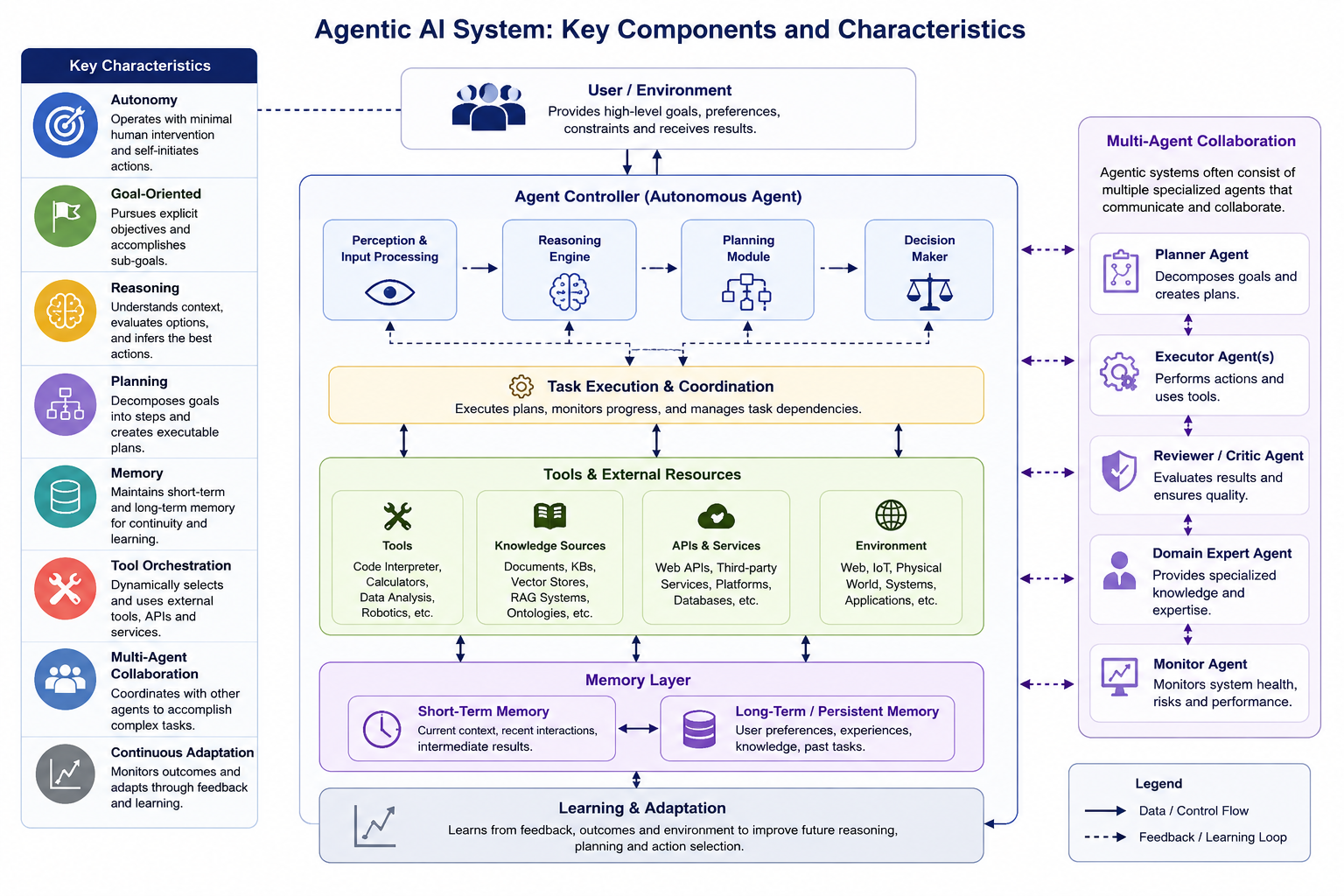}
    \caption{\textbf{Conceptual architecture and key characteristics of Agentic AI systems.} The figure illustrates the key components of Agentic AI, including autonomy, reasoning, planning, memory, tool orchestration, continuous adaptation, and multi-agent collaboration. Together, these capabilities enable agents to autonomously pursue goals, interact with external tools and environments, and coordinate complex tasks in dynamic settings.}\label{fig:AgTD_xtics}

\end{figure*}

These characteristics collectively distinguish Agentic AI from earlier AI paradigms. While Traditional AI focuses on task-specific prediction and Generative AI emphasizes content creation, Agentic AI integrates autonomy, reasoning, planning, memory, tool usage, and collaboration into unified systems capable of executing complex workflows in dynamic environments. As illustrated in Figure~\ref{fig:AgTD_xtics}, this evolution represents a transition from static, model-centric systems toward adaptive ecosystems of interacting agents. While these capabilities unlock significant opportunities across domains such as healthcare, finance, cybersecurity, and software engineering, they also introduce new forms of complexity, dependency, and emergent behavior that have important implications for technical debt management in agentic systems.

\subsubsection{Comparative Analysis of AI Paradigms}

\begin{figure*}[t]
    \centering
    \includegraphics[width=1.0\textwidth]{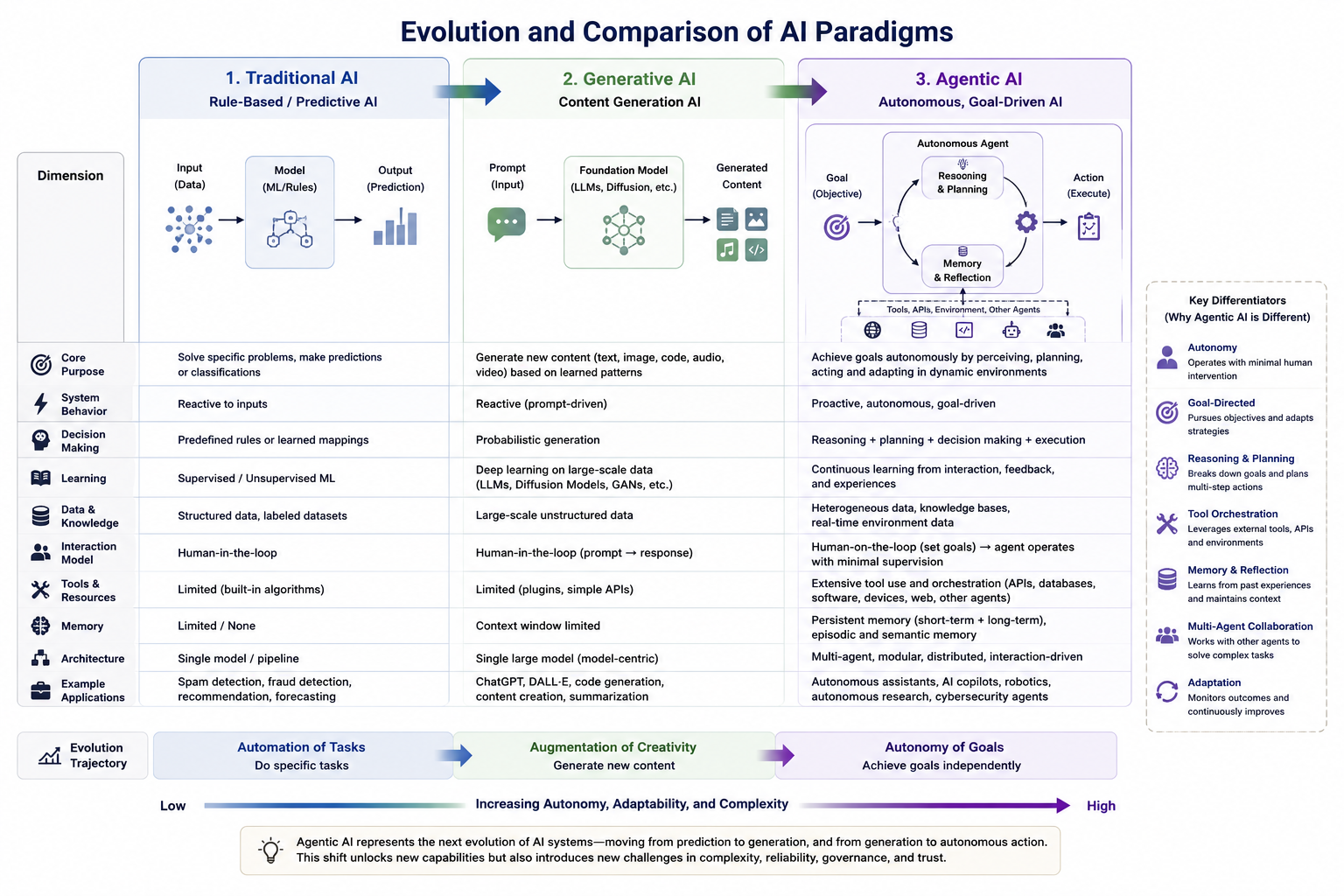}
    \caption{\textbf{Evolution and comparison of AI paradigms.} The figure illustrates the progression from Traditional AI to Generative AI and Agentic AI, highlighting key differences in autonomy, decision-making, learning, memory, interaction, and system architecture. It demonstrates the shift from task-specific prediction and automation, to content generation, and ultimately to autonomous, goal-directed systems capable of reasoning, planning, tool orchestration, memory management, and multi-agent collaboration.}\label{fig:AIParadigm_Comparisons}

\end{figure*}

To better contextualize these paradigms, Figure~\ref{fig:AIParadigm_Comparisons} illustrates the evolution of AI from Traditional AI to Generative AI and ultimately to Agentic AI, while Table~\ref{tab:ai_comparison} provides a detailed comparison across key dimensions, including system behavior, decision-making mechanisms, learning capabilities, interaction models, and architectural structure. Together, they highlight the fundamental shift from static, rule-based systems to dynamic, data-driven models, and ultimately to autonomous, goal-directed agentic systems. Traditional AI systems are largely deterministic and pipeline-oriented, operating under predefined logic with limited adaptability. Generative AI introduces probabilistic reasoning and content generation capabilities, enabling more flexible and creative interactions, yet it remains predominantly reactive and dependent on user prompts. In contrast, Agentic AI represents a paradigm shift toward autonomy and proactivity, where systems can independently plan, reason, and execute multi-step tasks while interacting with other agents, tools, and environments.

A key distinction lies in the level of system integration and coordination. While traditional and generative systems are typically model-centric, agentic systems are inherently distributed and multi-agent in nature, relying on coordinated interactions, shared memory, and external tool orchestration. As illustrated in Figure~\ref{fig:AgTD_xtics}, these systems integrate capabilities such as autonomy, reasoning, planning, memory, tool orchestration, continuous adaptation, and multi-agent collaboration within a unified architecture. This transition significantly increases system complexity but also unlocks new capabilities for solving long-horizon and real-world problems. Overall, while Generative AI extends the capabilities of traditional AI by enabling content creation and enhancing human-AI interaction, Agentic AI fundamentally transforms AI systems into autonomous, interactive, and adaptive entities capable of complex task execution, continuous learning, and collaborative problem-solving in dynamic environments.

\begin{table*}[h]
\centering
\caption{Comparison of Traditional AI, Generative AI, and Agentic AI}
\label{tab:ai_comparison}
\begin{tabular}{p{3cm} p{3.5cm} p{3.5cm} p{4cm}}
\toprule
\textbf{Attribute} & \textbf{Traditional AI} & \textbf{Generative AI} & \textbf{Agentic AI} \\
\midrule
Core Paradigm & Rule-based / Predictive & Content generation & Autonomous goal-driven systems \\
Behavior & Reactive & Reactive (prompt-driven) & Proactive and autonomous \\
Decision-Making & Predefined logic & Probabilistic generation & Planning + reasoning + execution \\
Learning & Supervised/unsupervised ML & Deep learning (LLMs, diffusion) & Continuous and adaptive learning \\
Interaction & Human-controlled & Human-in-the-loop & Multi-agent + environment interaction \\
System Structure & Pipeline-based & Model-centric & Multi-agent, distributed \\
Memory & Limited / static & Context window-based & Persistent and shared memory \\
Tool Usage & Minimal & Limited (plugins/APIs) & Extensive tool and API integration \\
Example Applications & Fraud detection, classification & ChatGPT, image generation & Autonomous assistants, AI agents, robotics \\
\bottomrule
\end{tabular}
\end{table*}

\subsection{Technical Debt: From Traditional Systems to AI}

The concept of Technical Debt (TD) was first introduced by Cunningham~\cite{cunningham1992wycash}, describing the trade-off between short-term development gains and long-term system maintainability. Technical debt arises when suboptimal design or implementation decisions are made to achieve rapid progress, leading to increased future costs in terms of maintenance, scalability, and system evolution~\cite{yang2023technical}. Traditional technical debt is typically associated with software engineering artifacts such as code quality, architecture, documentation, and testing. It is generally characterized as localized, being confined to specific components or modules; static, remaining relatively stable once introduced; predictable, with impacts that can be estimated and managed; and code-centric, primarily related to implementation and design decisions~\cite{brown2010managing}. However, the evolution of AI systems has introduced new forms of technical debt that extend beyond traditional software concerns~\cite{sculley2015hidden}.

\subsection{AI Technical Debt (AITD)}

AITD extends the notion of technical debt to machine learning and AI-enabled systems, encompassing issues related to data dependencies, model behavior, pipeline complexity, infrastructure integration, and operational processes~\cite{bogner2021characterizing}. Prior studies have shown that AI systems introduce unique sources of debt, including data quality and drift ~\cite{akgul2025aligning}, hidden feedback loops~\cite{khritankov2021hidden}, model entanglement~\cite{wang2023technical}, and evaluation limitations~\cite{tang2021empirical}. Unlike traditional technical debt, AITD is inherently data-centric, with a strong dependency on data quality and lifecycle; it involves complex multi-stage pipelines that introduce hidden dependencies; it exhibits non-deterministic behavior, where model outputs may vary across runs; and it is feedback-driven, as continuous learning introduces evolving system behavior~\cite{sklavenitis2025scoping}. As highlighted in the literature, many of these debts arise from the interaction between data, models, infrastructure, deployment pipelines, and human operational practices,  rather than purely from code-level issues, representing a shift from software-centric to system-centric technical debt~\cite{akgul2025aligning, bogner2021characterizing, sculley2015hidden}.

\subsection{Agentic Technical Debt (AgTD)}

Building on AITD~\cite{tukur2026aisafetysecuritytechnical}, the emergence of Agentic AI introduces a new class of technical debt-Agentic Technical Debt (AgTD). As established in this work, AgTD refers to technical debt that arises, accumulates, and propagates due to the autonomous, interactive, and adaptive nature of agentic systems (Please refer to Section~\ref{sec:agtd}). Agentic systems fundamentally differ from traditional AI systems in that they operate across multiple interacting agents, continuously adapt through feedback and learning, depend on shared memory and external tools, and exhibit emergent system-level behavior~\cite{moralles2026systematic}. These characteristics transform technical debt from a localized and static issue into a dynamic, distributed, and emergent phenomenon. As discussed in the manuscript, technical debt in agentic systems can propagate across agents, leading to coordination failures, cascading errors, and unintended behaviors.

\subsubsection{Comparative Perspective: TD vs AITD vs AgTD}

Table~\ref{tab:td_comparison} summarizes the key differences between traditional technical debt, AI technical debt, and agentic technical debt across key dimensions. The comparison highlights a clear evolution from component-level, deterministic, and manageable debt in traditional software systems to data- and model-driven complexities in AI systems, and ultimately to system-level, probabilistic, and emergent forms of debt in agentic AI environments. In particular, while traditional technical debt is largely confined to code and architectural decisions with relatively predictable impacts, AITD introduces additional challenges stemming from data dependencies, pipeline interactions, and non-deterministic model behavior. AgTD further amplifies these challenges by embedding debt within dynamic agent interactions, coordination mechanisms, and adaptive decision processes. As a result, technical debt transitions from a localized engineering concern to a distributed, evolving phenomenon that affects the overall behavior, reliability, and trustworthiness of the system. This progression underscores the increasing complexity of managing technical debt in modern AI systems and motivates the need for new models, tools, and governance strategies tailored to agentic environments.

\begin{table*}[h]
\centering
\caption{Comparison of Technical Debt Paradigms}
\label{tab:td_comparison}
\begin{tabular}{p{3cm} p{3.5cm} p{3.5cm} p{4cm}}
\toprule
\textbf{Attribute} & \textbf{Traditional TD} & \textbf{AITD} & \textbf{AgTD} \\
\midrule
Scope & Code/components & ML pipelines & Multi-agent ecosystems \\
Nature & Static & Semi-dynamic & Highly dynamic \\
Focus & Code quality & Data + models + pipelines & Interaction + behavior + coordination \\
Propagation & Limited/local & Pipeline-level & System-wide (cross-agent) \\
Behavior & Deterministic & Probabilistic & Emergent and adaptive \\
Complexity & Moderate & High & Very high \\
Traceability & High & Moderate & Low (due to interactions) \\
Risk Type & Maintainability & Model/data risks & Systemic and emergent risks \\
Examples & Code smells, poor architecture & Data debt, model bias & Coordination failures, memory corruption \\
\bottomrule
\end{tabular}
\end{table*}

\subsection{Summary}
In summary, AI has evolved from rule-based systems to generative models and, more recently, to agentic systems capable of autonomous and collaborative behavior. This evolution has fundamentally transformed the nature of technical debt—from static, code-centric issues to dynamic, system-level phenomena. While AITD captures the complexities introduced by data and models, AgTD further extends this concept to account for autonomy, interaction, and emergence in agentic ecosystems. This progression highlights the need for rethinking traditional software engineering and technical debt management approaches in the context of modern AI systems.

\section{Related Work}
\label{sec:relatedwork}

The rapid evolution of \textit{Agentic Artificial Intelligence (Agentic AI)} has led to a growing body of literature spanning conceptual foundations, system architectures, governance frameworks, and domain-specific applications. This section synthesizes existing work across four key dimensions: (i) conceptual and architectural foundations, (ii) systematic reviews and taxonomies, (iii) trust, risk, and governance, and (iv) domain-specific applications.

\subsection{Conceptual and Architectural Foundations}

Agentic AI is widely recognized as a paradigm shift from traditional and generative AI toward systems capable of autonomous, goal-directed behavior. Acharya et al~\cite{acharya2025agentic} define Agentic AI as autonomous systems designed to pursue complex goals with minimal human intervention, emphasizing adaptability, advanced decision-making, and self-sufficiency. Similarly, Pati~\cite{pati2025agentic} highlights that Agentic AI integrates autonomy, memory, goal-directed reasoning, and adaptive learning, enabling systems to operate proactively in dynamic environments. Beyond general definitions, Ali et al.,~\cite{abou2025agentic} introduces a novel dual-paradigm perspective, distinguishing agentic systems into symbolic/classical and neural/generative lineages. This work argues that paradigm selection is context-dependent, with symbolic approaches dominating safety-critical domains, while neural approaches excel in data-rich adaptive environments. Furthermore, Ali et al.~\cite{abou2025agentic} emphasizes the need for hybrid neuro-symbolic architectures to balance reliability and adaptability. Islam et al. ~\cite{islam2026rise}, extend this perspective by positioning Agentic AI as a socio-technical partner capable of integrating reasoning, planning, memory, and tool orchestration to achieve complex objectives with minimal human oversight. This evolution highlights the transition from reactive AI systems to proactive and collaborative intelligent agents.

\subsection{Systematic Reviews and Research Taxonomies}
Several systematic reviews have attempted to structure the fragmented landscape of Agentic AI research. Moralles et al~\cite{moralles2026systematic} present a comprehensive SLR that organizes the field into five major domains: memory cognition, networking systems, trust and safety, evaluation limits, and application use cases. Notably, it identifies a critical ``adaptability gap,'' revealing a lack of architectural frameworks capable of supporting long-term resilience and dynamic behavior in real-world environments. Complementing this, Hosseini and Zeilani~\cite{hosseini2025role} explore the role of Agentic AI in shaping intelligent organizational systems, identifying key attributes such as autonomy, reactivity, proactivity, and learning capability. It further highlights the transition from assistive (``copilot'') to autonomous (``autopilot'') systems and underscores the importance of hierarchical multi-agent architectures for coordination and scalability.
Raheem and Hossain~\cite{raheem2025agentic} provide a broader perspective on the opportunities and challenges of Agentic AI, emphasizing its ability to enhance efficiency, scalability, and decision-making in organizational contexts, while also raising concerns related to safety, reliability, and accountability.

\subsection{Trust, Risk, Security, and Governance}
As Agentic AI systems become more autonomous and distributed, trust and governance have emerged as critical research areas. Raza et al.,~\cite{RAZA202671} introduce a comprehensive Trust, Risk, and Security Management (TRiSM) framework tailored for LLM-based Agentic Multi-Agent Systems (AMAS). This work proposes a structured approach covering explainability, ModelOps, security, privacy, and lifecycle governance.
Importantly, RW2 also introduces novel evaluation metrics, such as the Component Synergy Score (CSS) and Tool Utilization Efficacy (TUE), to quantify inter-agent coordination and tool usage effectiveness—an area largely overlooked in earlier studies.
From a security perspective, Rashid et al.,~\cite{rashid2025securing} systematically analyze threats and vulnerabilities in Agentic AI systems, including prompt-based adversarial attacks, coordination failures, and system-level risks. It highlights ongoing efforts such as the OWASP Agentic AI Top 10 initiative, emphasizing the need for standardized security frameworks.
Across these studies, a common challenge emerges: ensuring safety, transparency, and accountability in increasingly autonomous and complex agentic ecosystems.

\subsection{Domain-Specific Applications}
Agentic AI has been applied across multiple domains, demonstrating its versatility and transformative potential.
In healthcare, Collaco et al. \cite{collaco2026role}, present a scoping review showing that agentic systems can autonomously achieve clinical goals such as diagnosis, treatment planning, alert generation, and workflow optimization. However, the study highlights that most implementations remain exploratory and lack robust clinical validation.
In smart grids, Kiasari and Aly~\cite{kiasari2026agentic} provide a comprehensive review of Agentic AI for cyber–physical energy systems, demonstrating its ability to perform real-time perception, multi-step planning, adaptive coordination, and autonomous control. Applications include voltage control, fault detection, distributed energy resource coordination, and grid restoration, emphasizing reliability and safety in critical infrastructures. 
In education, Kostopoulos et al.,~\cite{kostopoulos2025agentic} explore Agentic AI as autonomous learning companions capable of delivering personalized instruction, adaptive feedback, and dynamic interaction with learners, highlighting both pedagogical benefits and ethical concerns.
Across domains, Agentic AI systems consistently demonstrate autonomy, proactivity, and multi-step reasoning, while also exposing challenges related to validation, safety, and real-world deployment.

\subsection{Research Gaps and Limitations}

Despite recent advances, key gaps remain from an AgTD perspective:

\begin{itemize}

    \item \textbf{Unaddressed Sources of Agentic Technical Debt:} Existing studies (e.g., \cite{moralles2026systematic}) do not adequately address debt arising from autonomous decision-making, dynamic task orchestration, and inter-agent dependencies.
    

    \item \textbf{Lack of Unified Modeling for AgTD Propagation:} Current research lacks systematic approaches to model how technical debt propagates across agent interactions, coordination workflows, and emergent system behaviors, which serve as key pathways through which AgTD accumulates and amplifies at the system level.
    

    \item \textbf{Insufficient Governance for Autonomous Systems:} While frameworks such as TRiSM~\cite{RAZA202671} provide foundations for trust, risk, and security management, they do not fully address accountability, traceability, and control challenges in agentic ecosystems. These governance gaps can contribute to the accumulation of Agentic Technical Debt (AgTD) by limiting the visibility, monitoring, and remediation of debt arising from autonomous decisions, agent interactions, and evolving system behaviors.
    

    \item \textbf{Underexplored Security and Coordination Debt:} Technical debt arising from inadequate security controls, coordination mechanisms, and agent interaction protocols remains underexplored in agentic systems. Such debt can increase susceptibility to risks such as adversarial manipulation and coordination failures~\cite{rashid2025securing}, limiting proactive mitigation and system resilience.

    \item \textbf{Underexplored Sustainability Technical Debt:} Although recent studies have extensively discussed trustworthiness, safety, security, and governance in Agentic AI~\cite{raza2025trism, raheem2025agentic, leo2026threat, adabara2025trustworthy, qi2026towards}, the sustainability implications of AgTD remain largely unexplored. Agentic AI systems are inherently computationally intensive due to continuous reasoning, autonomous planning, persistent memory management, repeated inference, multi-agent communication, and frequent tool orchestration~\cite{abou2026agentic}. Design deficiencies and inefficient workflow structures may therefore accumulate as \emph{SusTD}, leading to excessive energy consumption, increased operational costs, inefficient resource utilization, and a larger carbon footprint throughout the system lifecycle. Despite the growing emphasis on Green AI and sustainable software engineering~\cite{cruz2025innovating, alloghani2023architecting}, sustainability has received limited attention as a distinct manifestation of AgTD, highlighting an important research gap for future investigation.

\end{itemize}

\subsection{Positioning of This Work}
While prior studies provide comprehensive insights into the concepts, architectures, applications, and governance of Agentic AI, they largely overlook its implications for software quality, long-term system sustainability, and software engineering practices. In particular, existing research does not explicitly conceptualize how the autonomous, adaptive, collaborative, and persistent nature of agentic systems gives rise to new forms of technical debt that extend beyond both traditional technical debt and AI Technical Debt (AITD).

This paper addresses this gap by introducing \textit{Agentic Technical Debt (AgTD)} as a foundational software engineering construct for understanding and managing technical debt in Agentic AI systems. Building upon a previously validated taxonomy of 31 AITDs~\cite{tukur2026aisafetysecuritytechnical}, this work presents a theory-informed methodology for systematically transforming established AITDs into their corresponding agentic manifestations through transformation semantics. Specifically, the paper provides (i) a formal definition of AgTD, (ii) a systematic mapping of 31 AITDs to their agentic counterparts, (iii) a taxonomy-driven characterization of AgTD grounded in direct, contextual, and expansion transformations, and (iv) an analysis of the unique debt mechanisms introduced by autonomy, multi-agent coordination, tool integration, persistent memory, and adaptive reasoning.

Furthermore, this study extends beyond technical characterization by examining the broader implications of AgTD for AI Trust, Risk, and Security Management (TRiSM) and long-term software sustainability. We demonstrate how AgTD contributes to trust degradation, systemic risk amplification, governance complexity, explainability challenges, and increased security vulnerabilities, while also introducing emerging forms of Sustainability Technical Debt (SusTD) associated with resource-intensive reasoning, persistent context management, and large-scale autonomous agent orchestration.

By shifting the focus from \textit{agent capabilities} to \textit{agent-induced system liabilities}, this work provides a unified conceptual foundation for understanding, analyzing, and managing software quality, trustworthiness, security, sustainability, and long-term evolution in Agentic AI systems. More broadly, it establishes AgTD as a new software engineering perspective that complements existing research on Agentic AI and lays the foundation for future advances in runtime monitoring, governance, measurement, validation, refactoring, and technical debt mitigation for autonomous multi-agent ecosystems.

\section{Methodology}
\label{sec:methodology}

\begin{figure*}[t]
    \centering
    \includegraphics[width=1.0\textwidth]{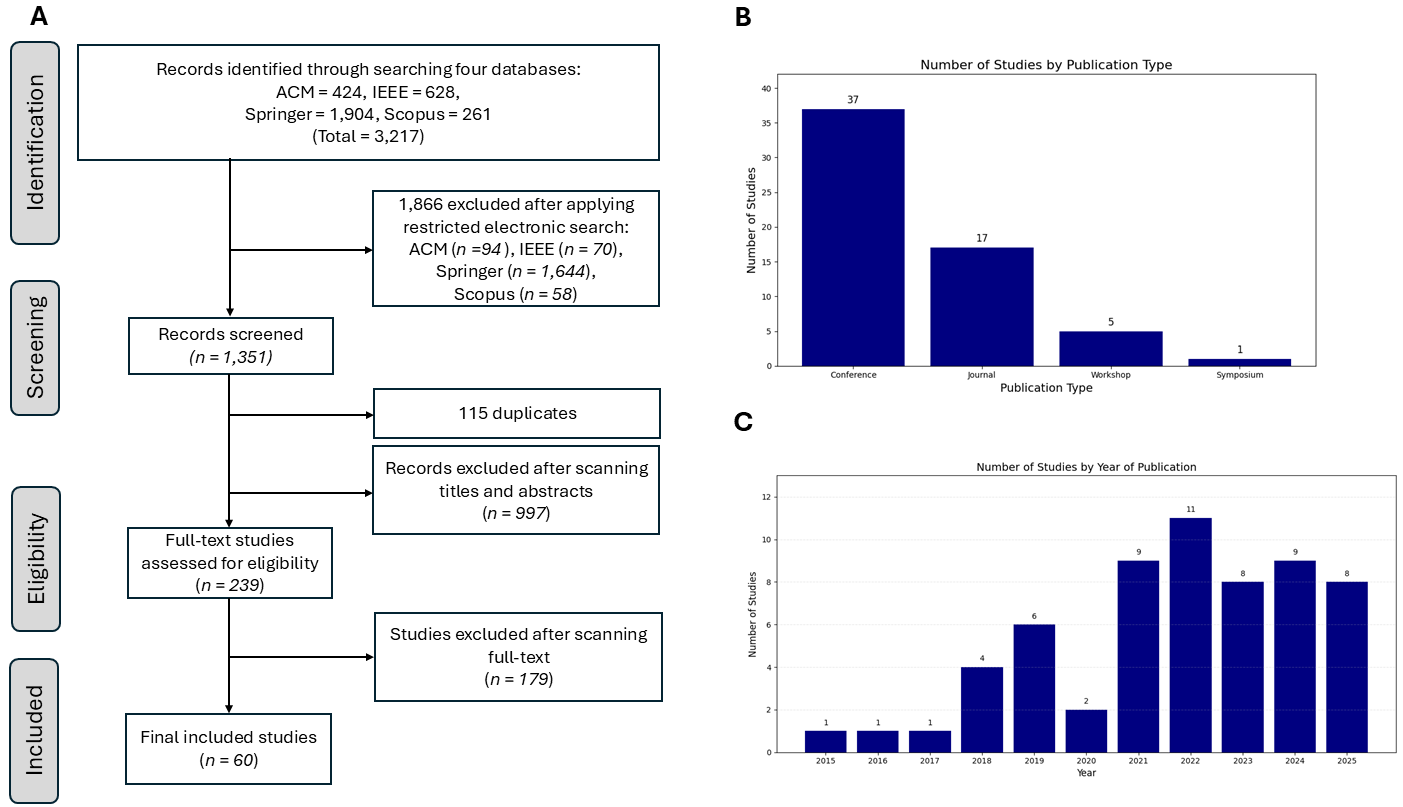}
    
    \caption{Methodology charts: (A) PRISMA chart of the included studies; (B) publication type of the
selected papers. (C) The distribution of studies over the years. 
}\label{fig:prismachart}

\end{figure*}

This study adopts a \textit{multi-phase, theory-informed research methodology} that builds upon our prior systematic scoping review of AI Technical Debt (AITD)~\cite{tukur2026aisafetysecuritytechnical} to develop the conceptual foundation, reinterpret existing AI technical debts, and analyze their manifestations in the context of \textit{Agentic Technical Debt (AgTD)}.
Unlike traditional SLR-based studies, this work does not aim to re-identify technical debts, but rather to \textit{reinterpret and transform} existing AITDs within the context of Agentic AI systems, characterized by autonomy, multi-agent coordination, tool integration, and persistent memory. As illustrated in Figure~\ref{fig:prismachart}.A, the methodology is grounded in the PRISMA-ScR framework \cite{tricco2018prisma} for the initial AITD identification, followed by a structured qualitative mapping and abstraction process tailored to Agentic AI. Overall, this methodology provides a rigorous and extensible framework for systematically transforming existing knowledge on AI Technical Debt into a new conceptual foundation for understanding, categorising, and analysing technical debt in next-generation Agentic AI systems.

\subsection{Phase 1: Source Taxonomy (AITD Extraction)}

The foundation of this work is our previously conducted PRISMA-compliant scoping review that systematically identified \textbf{31 AITDs} across \textbf{seven root-cause categories}~\cite{tukur2026aisafetysecuritytechnical}. The review followed a rigorous multi-stage process, including database search, screening, and full-text analysis, resulting in 60 primary studies (see Table~\ref{tab:venue-distribution}).

\begin{figure*}[t]
    \centering
    \includegraphics[width=1.0\textwidth]{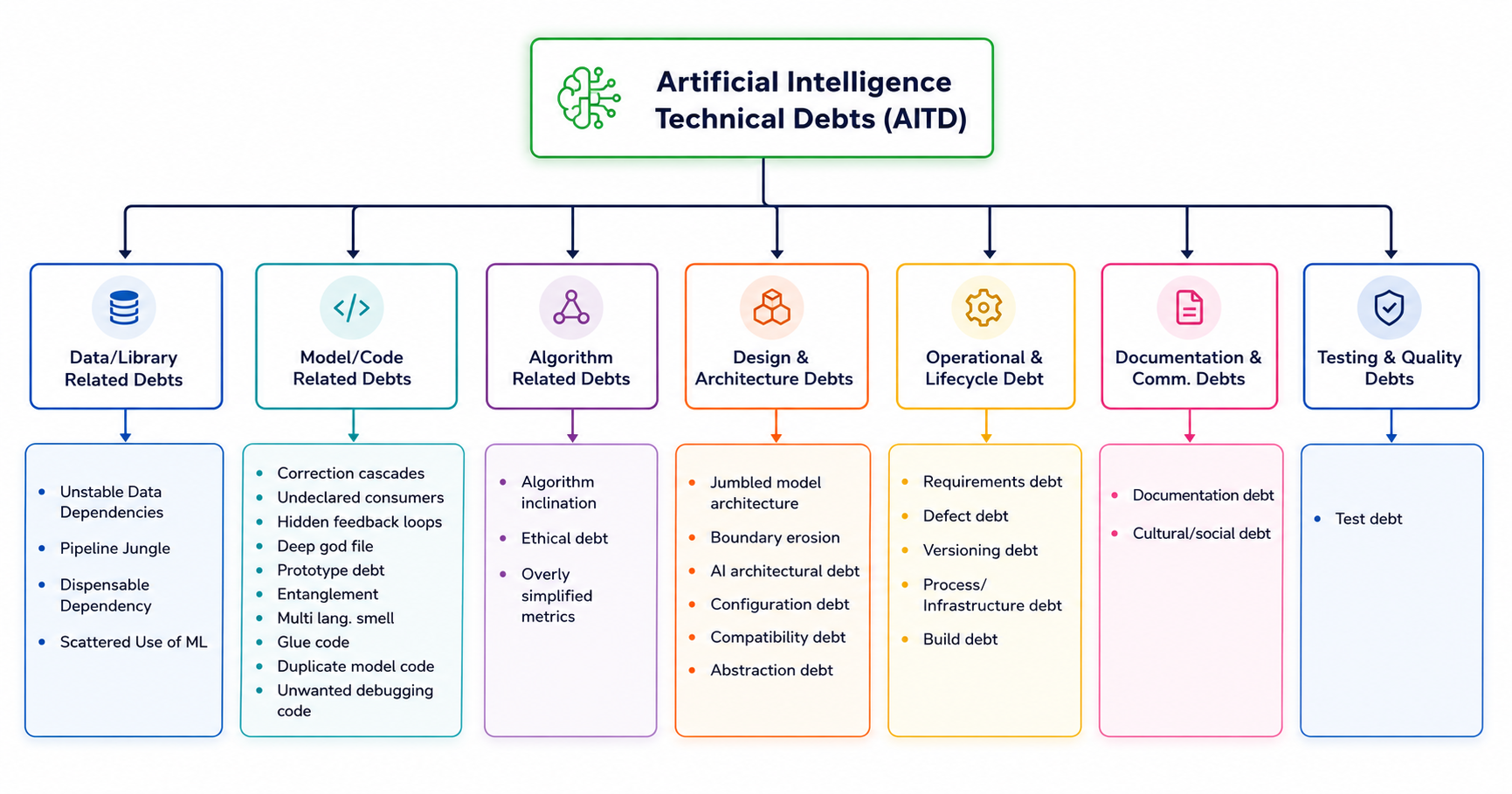}
    \caption{AITD taxonomy showing seven main categories-each divided into subcategories with corresponding debt types.}\label{fig:AITD_Taxonomy}

\end{figure*}

This phase ensures that the identified AITDs are:
\begin{itemize}
    \item Empirically grounded in software engineering literature
    \item Representative of AI-enabled systems (2015--2025)
    \item Organized into a validated root-cause taxonomy (see Fig.~\ref{fig:AITD_Taxonomy}, and Table~\ref{tab:AITDranking}).
\end{itemize}

Rather than re-performing the review, this study \textit{leverages the validated AITD taxonomy as a baseline knowledge model} for further analysis.

\begin{table*}[ht]
\small
\centering
\caption{Distribution of Selected Studies by Venue}
\begin{tabular}{|c|p{12cm}|c|}
\hline
\textbf{S/N} & \textbf{Venue} & \textbf{Count} \\
\hline
1 & International Conference on AI Engineering: Software Engineering for AI (CAIN) & 6 \\
2 & Empirical Software Engineering & 5 \\
3 & Euromicro Conference on Software Engineering and Advanced Applications (SEAA) & 4 \\
4 & International Conference on Software Engineering (ICSE) & 4 \\
5 & ACM/IEEE International Conference on Technical Debt (TechDebt) & 4 \\
6 & ACM Transactions on Software Engineering and Methodology & 3 \\
7 & IEEE International Conference on Software Maintenance and Evolution (ICSME) & 3 \\
8 & IEEE Transactions on Software Engineering & 2 \\
9 & Software Quality: Future Perspectives on Software Engineering Quality (SWQD) & 2 \\
10 & AI and Ethics & 2 \\
11 & International Conference on Mining Software Repositories (MSR) & 2 \\
12 & Journal of Systems and Software & 1 \\
13 & ACM Conference on Equity and Access in Algorithms, Mechanisms, and Optimization & 1 \\
14 & ACM Conference on Fairness, Accountability, and Transparency & 1 \\
15 & ACM Joint European Software Engineering Conference and Symposium on the Foundations of Software Engineering & 1 \\
16 & ACM SIGMOD Record & 1 \\
17 & ACM/IEEE Workshop on AI Engineering-Software Engineering for AI (WAIN) & 1 \\
18 & Advances in Neural Information Processing Systems & 1 \\
19 & IEEE International Conference on Big Data (Big Data) & 1 \\
20 & Information and Software Technology & 1 \\
21 & International Conference on Automated Software Engineering (ASE) & 1 \\
22 & International Conference on Green Computing and Internet of Things (ICGCIoT) & 1 \\
23 & International Conference on Industrial Informatics (INDIN) & 1 \\
24 & International Conference on Product-Focused Software Process Improvement & 1 \\
25 & International Conference on Software Architecture Companion (ICSA-C) & 1 \\
26 & International Requirements Engineering Conference Workshops & 1 \\
27 & International Workshop on Empirical Software Engineering in Practice (IWESEP) & 1 \\
28 & Workshop on Machine Learning and Systems & 1 \\
29 & World Wide Web Conference & 1 \\
30 & International Conference on Emerging Technologies and Computing (ICETC) & 1 \\
31 & Brazilian Symposium on Software Components, Architectures, and Reuse & 1 \\
32 & International Conference on Program Comprehension (ICPC) & 1 \\
33 & IEEE Access & 1 \\
34 & IEEE Annual Computing and Communication Workshop and Conference (CCWC) & 1 \\
\hline
\multicolumn{2}{|c|}{\textbf{Total}} & \textbf{60} \\
\hline
\end{tabular}
\label{tab:venue-distribution}
\end{table*}

\begin{table*}[ht]
\centering
\caption{Overview of AITDs and their categorization as part of the systematic review}
\label{tab:AITDranking}
\begin{tabular}{c l c c l}
\toprule
\textbf{\#} & \textbf{AITD} & \textbf{Freq(n=60)} & \textbf{\%} &
\textbf{Category}\\ \hline

1  & Data Debt/Unstable Data Dependencies  & 27 & 45.00 & Data/Library Related Debts \\ \hline
2  & Glue Code (GC) & 18 & 30.00 &Model/Code Related Debts \\ \hline
3  & Test Debt & 17 & 28.33 & Testing \& Quality Assurance Debts\\ \hline
4  & Documentation Debt & 15 & 25.33 & Documentation \& Communication Debts \\ \hline
5  & Requirement Debt & 14 & 23.33 & Operational \& Lifecycle Debts\\ \hline

6  & AI Architectural Debt & 13 & 21.67 & Design \& Architecture Debts \\ \hline

7  & Configuration Debt & 12 & 20.00 & Design \& Architecture Debts\\ \hline
8 & Dead Experimental Code Paths/Prototype Debt & 11 & 18.33 & Model/Code Related Debts \\ \hline

9 & Algorithm Debt/Inclination/Human Bias & 10 & 16.67 & Algorithm Related Debts \\ \hline

10 & Duplicate Model Code & 10 & 16.67  & Model/Code Related Debts \\ \hline

11 & Cultural/People/Social Debt & 10 & 16.67 & Documentation \& Communication Debts \\ \hline
12  & Pipeline Jungle & 9 & 15.00  & Data/Library Related Debts \\ \hline
13 & Jumbled Model Architecture (JMA) & 9 & 15.00 & Design \& Architecture Debts \\ \hline

14 & Hidden Feedback Loops & 9 & 15.00 & Model/Code Related Debts \\ \hline

15  & Defect Debt & 9 & 15.00 & Operational \& Lifecycle Debts \\ \hline

16 & Multiple Language Smells (MLS) & 7  & 11.67 & Model/Code Related Debts\\ \hline

17 & Undeclared consumers & 6  & 10.00 & Model/Code Related Debts \\ \hline
18 & Ethical Debt & 6  & 10.00 & Algorithm Related Debts\\ \hline

19 & Unwanted Debugging Code (UDC) & 5  & 8.33 &Model/Code Related Debts\\ \hline

20 & Versioning Debt & 5  & 8.33 & Operational \& Lifecycle Debts\\ \hline

21 & Correction Cascades (CC) & 4  & 6.67 & Model/Code Related Debts \\ \hline
22 & Entanglement & 4  & 6.67 & Model/Code Related Debts\\ \hline

23 & Compatibility Debt & 4  & 6.67 & Design \& Architecture Debts \\ \hline

24 & Abstraction Debt & 4  & 6.67 & Design \& Architecture Debts \\ \hline

25 & Dispensable Dependency & 3  & 5.00 & Data/Library Related Debts \\ \hline
26 & Boundary Erosion & 3  & 5.00 & Design \& Architecture Debts \\ \hline

27 & Overly Simplified Metrics & 3  & 5.00 & Algorithm Related Debts \\ 
\hline

28 & Process/Infrastructure Debt & 3 & 5.00 & Operational \& Lifecycle Debts \\ \hline
29 & Build Debt & 3  & 5.00 & Operational \& Lifecycle Debts\\ \hline
30 & Scattered Use of ML Libraries (SML) & 2  & 3.33 & Data/Library Related Debts\\ \hline

31 & Deep God File (DG) & 2  & 3.33 & Model/Code Related Debts \\

\bottomrule
\end{tabular}

\end{table*}

\subsection{Phase 2: Agentic Contextualization}

To extend AITDs into the Agentic AI paradigm, we first established a conceptual model of Agentic AI systems based on recent literature (Section~\ref{sec:relatedwork}). These systems are characterized by:

\begin{itemize}
    \item Autonomous goal-directed behavior
    \item Multi-agent coordination and interaction
    \item Integration with external tools and APIs
    \item Persistent memory and feedback loops
\end{itemize}

Each AITD was then reinterpreted under these characteristics to identify how its manifestation changes in \textit{dynamic, distributed, and interaction-driven environments}. This step enables the transition from static, pipeline-based assumptions to agentic system behavior.

\subsection{Phase 3: Mapping AITDs to Agentic Manifestations}

The core contribution of this work lies in a \textit{systematic mapping process} that transforms each of the 31 AITDs into its corresponding \textit{Agentic Technical Debt manifestation (AgTD)}.

The mapping was conducted through iterative qualitative analysis using the following transformation semantics:

\begin{itemize}
    \item \textbf{Direct Transformation:} AITD retains its structure but appears in agentic workflows (e.g., Glue Code $\rightarrow$ agent orchestration logic)
    \item \textbf{Contextual Transformation:} AITD changes due to agent autonomy and interaction (e.g., Hidden Feedback Loops $\rightarrow$ self-reinforcing agent loops)
    \item \textbf{Manifestation Expansion:} AITD evolves into system-level emergent behavior (e.g., Entanglement $\rightarrow$ multi-agent dependency networks)
\end{itemize}

\begin{figure*}[t]
    \centering
    \includegraphics[width=1.0\textwidth]{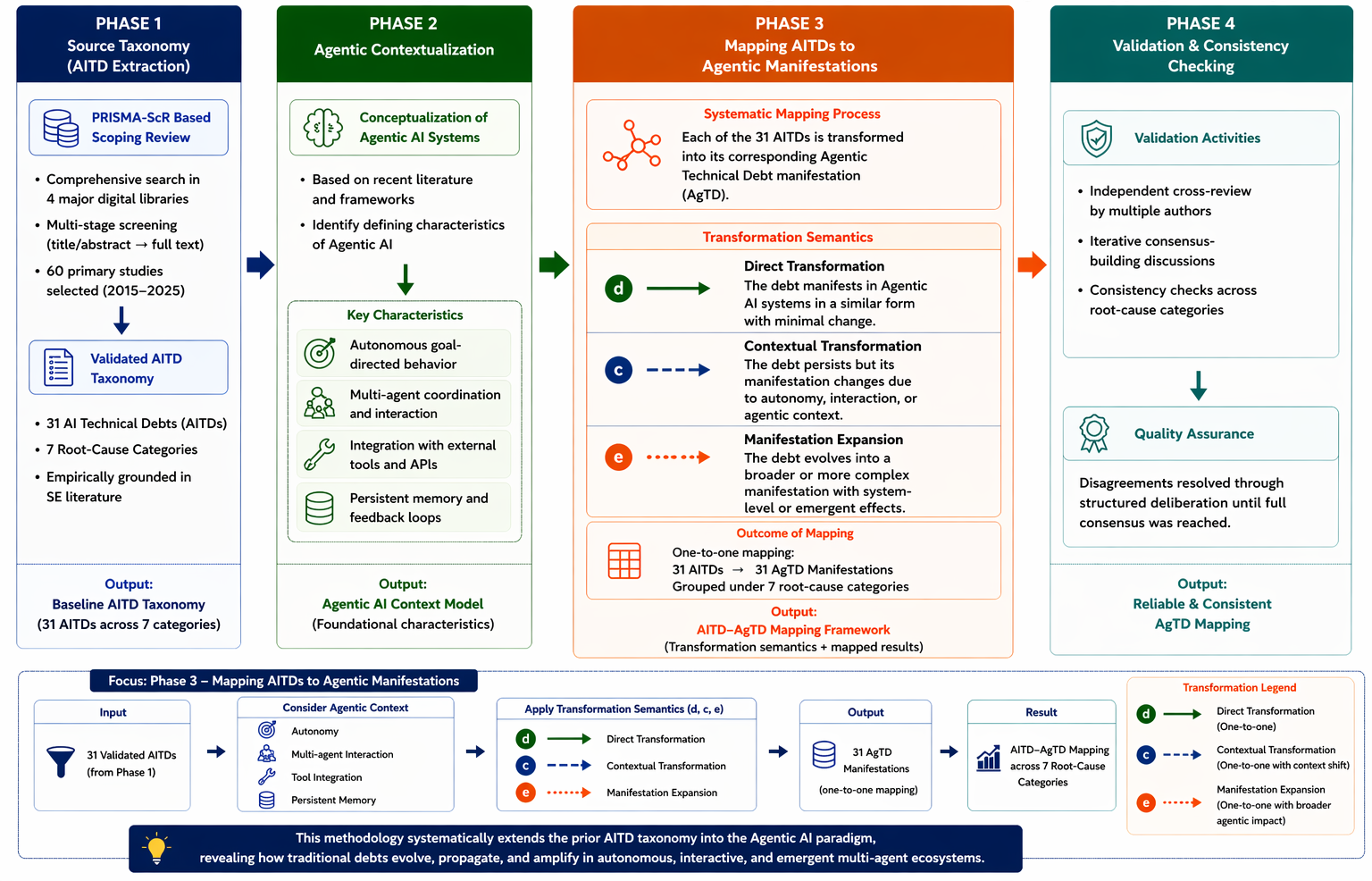}
    \caption{Overview of the methodology for mapping AITD to AgTD. The process includes four phases: (1) extraction of a validated AITD taxonomy via PRISMA-ScR, (2) contextualization within Agentic AI characteristics, (3) systematic mapping of AITDs to AgTD manifestations using transformation semantics—direct (d), contextual (c), and expansion (e), and (4) validation through cross-review and consistency checks. The lower panel details the Phase 3 mapping workflow and resulting one-to-one AgTD mapping across seven root-cause categories.}\label{fig:AITD_AgTD_Methodology}

\end{figure*}

This process resulted in a one-to-one mapping between AITDs and their agentic manifestations, as illustrated in the mapping framework (Figure~\ref{fig:AITD_AgTD_Methodology}).

\begin{figure*}[t]
    \centering
    \includegraphics[width=1.0\textwidth]{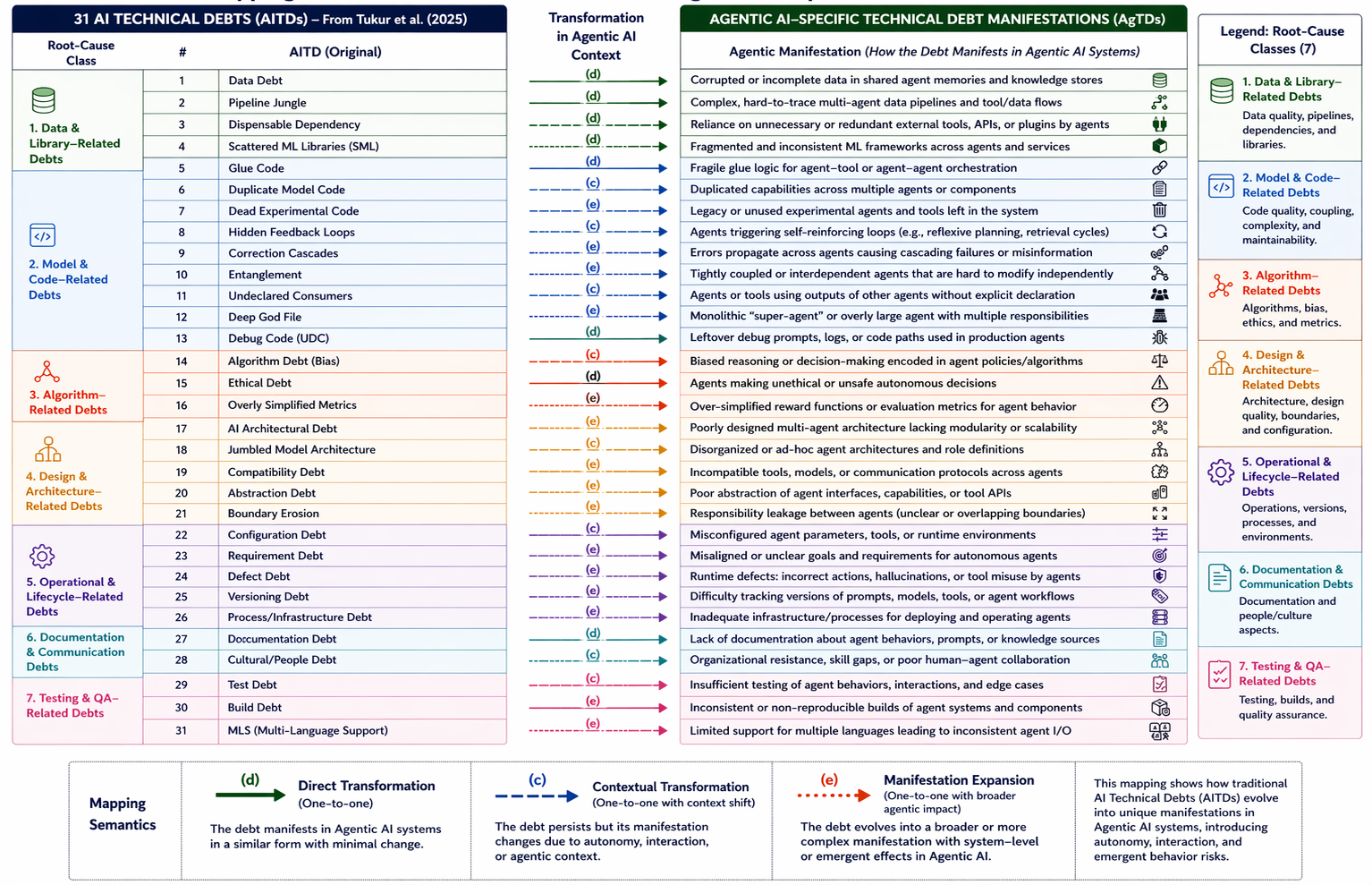}
    \caption{\textbf{Mapping of 31 AITDs to Agentic AI–specific Technical Debt Manifestations (AgTDs).} The figure illustrates how traditional AITDs, organized across seven root-cause categories, are systematically transformed within Agentic AI systems characterized by autonomy, multi-agent coordination, tool integration, and persistent memory. Each AITD is mapped to its corresponding agentic manifestation, highlighting three transformation semantics: (i) direct transformation, where the debt persists in agentic workflows; (ii) contextual transformation, where the debt adapts to agent interaction and autonomy; and (iii) manifestation expansion, where the debt evolves into system-level, emergent behaviors. The mapping reveals how component-level technical debts in conventional AI systems propagate into dynamic, distributed, and interaction-driven risks in agentic environments, including memory corruption, orchestration fragility, cascading failures, and coordination breakdowns. This figure serves as the core analytical framework for understanding the evolution of technical debt in Agentic AI systems.}\label{fig:AITD_to_AgTD_Mapping}

\end{figure*}









\subsection{Phase 4: Validation and Consistency Checking}

To enhance the reliability and validity of the mapping process, all identified Agentic Technical Debts (AgTDs) and their corresponding root-cause classifications underwent a structured validation process. Initially, the mappings were independently reviewed by multiple authors to assess their conceptual accuracy, completeness, and alignment with the original AI Technical Debt (AITD) definitions. Reviewers evaluated whether each proposed agentic manifestation preserved the fundamental characteristics of the source AITD while appropriately reflecting the unique properties of Agentic AI systems, such as autonomy, reasoning, memory, tool orchestration, and multi-agent collaboration.

Following the independent review, discrepancies and ambiguous cases were discussed through iterative consensus-building sessions. During these discussions, authors examined the rationale for each mapping, compared alternative interpretations, and refined descriptions where necessary. A mapping was accepted only when consensus was reached among the reviewers.

To ensure consistency across the taxonomy, additional cross-category checks were performed to verify that similar manifestations were classified uniformly and that no mapping was duplicated across multiple root-cause categories without justification. Particular attention was given to maintaining clear distinctions between related debt types while preserving traceability to the original AITD taxonomy.

This validation process ensured that each mapping remained both (i) faithful to the original AITD definition and intent, and (ii) consistent with the operational characteristics and behaviors of Agentic AI systems.

\subsection{Summary of Methodological Contribution}

Overall, this methodology extends traditional SLR-based AITD analysis into the domain of Agentic AI through a structured, multi-phase approach. It enables:

\begin{itemize}
    \item Systematic reinterpretation of existing technical debts
    \item Identification of agentic-specific manifestations
    \item Preservation of traceability from original AITDs to AgTD
\end{itemize}

This approach provides a rigorous foundation for analyzing how technical debt evolves in autonomous, multi-agent systems.

\section{Agentic Technical Debt (AgTD): Definition, Mapping, and Analysis}
\label{sec:agtd}

\subsection{Formal Definition of Agentic Technical Debt}

Building upon the classical notion of Technical Debt (TD)~\cite{cunningham1992wycash} and its subsequent extension to AI systems (AITD)~\cite{bogner2021characterizing, tukur2026aisafetysecuritytechnical}, we introduce \textit{AgTD} as a new paradigm that captures the evolving realities of modern Agentic AI systems. \textit{AgTD refers to the forms of technical debt that arise, accumulate, propagate, and amplify due to the autonomous, interactive, and adaptive behavior of agentic AI systems, particularly in environments characterized by multi-agent coordination, tool integration, and persistent memory.} 
In contrast to traditional technical debt, which primarily originates from software artifacts such as code, architecture, design decisions, and development processes and is often relatively predictable in its manifestation and impact~\cite{ernst2021technical}, AgTD emerges in systems that are inherently dynamic, distributed, and behavior-driven. While traditional technical debt can also evolve and propagate across components, AgTD is further shaped by autonomous decision-making, agent interactions, adaptive behaviors, tool orchestration, and multi-agent coordination. Consequently, debt is no longer confined to software artifacts alone but can also become embedded within the interactions, decisions, and evolving states of agents operating in complex environments.
Consistent with classical technical debt theory~\cite{ernst2021technical}, AgTD refers to the underlying design, implementation, architectural, or governance compromises that create future costs, rather than the resulting failures or system-level consequences that emerge from those compromises.


Specifically, AgTD exhibits several characteristics that become particularly pronounced in agentic AI systems. First, it demonstrates \textit{dynamic evolution}, whereby debt can continuously change as agents learn, adapt, and interact with internal and external environments. Second, it manifests in a highly \textit{distributed manner}, spanning multiple agents, tools, memory stores, and system layers, often with diffuse ownership and complex dependencies. Third, it can contribute to \textit{emergent behaviors}, where unintended system-level outcomes arise from interactions among autonomous agents rather than from individual components alone. Fourth, AgTD exhibits strong \textit{propagation effects}, whereby errors, inconsistencies, or misalignments can spread through agent interactions and coordinated workflows. Finally, it increases \textit{opacity}, as autonomous decision-making, reasoning processes, and complex agent interactions make it more difficult to trace root causes, explain behaviors, and ensure accountability.



Collectively, these characteristics extend technical debt theory beyond a component-centric perspective focused on code and architecture toward a runtime-centric paradigm in which debt emerges and evolves through autonomous decisions, agent interactions, and continuous adaptation. Unlike traditional systems, where debt is often associated with relatively stable software artifacts, agentic systems exhibit a high degree of runtime fluidity, with behaviors, dependencies, and execution pathways continuously changing during operation. This shift necessitates rethinking how technical debt is identified, measured, and managed~\cite{ernst2021technical}, moving toward continuous monitoring, runtime validation, and governance-aware strategies tailored to autonomous, adaptive, and multi-agent ecosystems.

\subsection{Systematic Mapping of AITDs to Agentic Manifestations}

To operationalize the concept of AgTD, we systematically map the 31 AITDs identified in prior work to their corresponding agentic manifestations (AgTDs), as illustrated in Figure~\ref{fig:AITD_to_AgTD_Mapping}. This mapping serves as a structured bridge between traditional AI-based technical debt and its evolved forms in agentic environments, enabling a clear and traceable transformation from component-level issues to system-level behaviors.

The mapping is designed to preserve several critical properties. First, it ensures \textit{traceability}, such that each AgTD can be directly linked to its originating AITD, maintaining conceptual continuity and interpretability. Second, it enforces \textit{taxonomy consistency}, where all mapped debts remain aligned with the original seven root-cause categories, thereby preserving the underlying theoretical structure of technical debt classification. Third, it maintains a strict \textit{one-to-one correspondence}, ensuring that each of the 31 AITDs is systematically transformed into a unique agentic manifestation, avoiding ambiguity or overlap in the mapping process.


Beyond structural consistency, the mapping provides important analytical insights into how technical debt evolves in Agentic AI systems. Specifically, it reveals a shift from isolated design and implementation concerns toward runtime, interaction-driven, and system-level consequences. In agentic environments, traditional debts are no longer confined to individual modules or pipelines; instead, their effects can propagate through complex agent interactions, shared memory spaces, and continuous feedback loops. As a result, debt-inducing decisions related to data management, coordination mechanisms, or agent orchestration may manifest as memory inconsistencies across agents, orchestration fragility, or cascading failures that propagate through agent networks, increasing the long-term cost and complexity of remediation.

Furthermore, the mapping highlights how certain debts escalate into higher-order risks, including \textit{unsafe autonomous decision-making}, \textit{emergent coordination breakdowns}, and \textit{loss of system transparency}. These transformations underscore the increasing complexity and unpredictability of agentic systems, where the interaction between autonomy, adaptation, and distributed execution amplifies both the visibility and impact of technical debt. Overall, this systematic mapping not only provides a rigorous foundation for defining AgTD but also offers a comprehensive lens for understanding how technical debt manifests, evolves, and propagates in modern Agentic AI systems.

\subsection{Transformation Semantics}

The transformation from AITD to AgTD follows three distinct semantics (Figure~\ref{fig:AITD_to_AgTD_Mapping}), which collectively describe how traditional technical debts evolve in the context of Agentic AI systems. These semantics provide a conceptual foundation for understanding whether a debt is preserved, adapted, or fundamentally transformed when moving from static, pipeline-based AI systems to dynamic, multi-agent environments characterized by autonomy, interaction, and continuous adaptation.

\subsubsection{Direct Transformation (d)}

In direct transformation, the technical debt persists in agentic systems with minimal structural modification, retaining its original form while becoming embedded within agent workflows, tool integrations, and execution pipelines. In this case, the underlying cause and nature of the debt remain largely unchanged; however, its operational context shifts from isolated components to distributed agentic environments. As a result, although the debt does not fundamentally evolve in structure, its impact may be amplified due to the scale, frequency, and interconnected nature of agent interactions. 

Direct transformations typically occur for debts that are inherently structural or implementation-driven, such as data inconsistencies, redundant dependencies, or orchestration logic. When these debts are carried into agentic systems, they become tightly coupled with agent execution flows, shared memory systems, and tool invocation mechanisms. Consequently, even minor inefficiencies or inconsistencies can propagate across multiple agents, increasing their visibility and operational impact without altering their core definition. 

To improve traceability and explain the rationale behind each transformation, Table~\ref{tab:direct_transformation} summarizes the original AITDs, their definitions, the mechanisms through which they manifest in Agentic AI systems, and their resulting AgTD manifestations. The table illustrates how these debts are directly inherited from traditional AI systems while becoming embedded within agent interactions, shared memory, tool orchestration, and autonomous decision-making processes. 

This semantic category highlights that not all technical debts undergo conceptual transformation in Agentic AI; instead, some are directly inherited from traditional AI systems but operate under more complex and demanding conditions. Understanding direct transformation is therefore essential for identifying legacy debt patterns that persist unchanged, yet become more critical in agentic settings due to increased system scale, autonomy, and interaction density.

\textbf{AgTDs under Direct Transformation:}
\begin{enumerate}
    \item Data Debt $\rightarrow$ corrupted or incomplete shared agent memory
    \item Pipeline Jungle $\rightarrow$ complex multi-agent data pipelines
    \item Dispensable Dependency $\rightarrow$ unnecessary tool/API reliance
    \item Scattered ML Libraries $\rightarrow$ fragmented agent frameworks
    \item Glue Code $\rightarrow$ fragile agent orchestration logic
    \item Debug Code (UDC) $\rightarrow$ leftover debug prompts/logs in agents
    \item Ethical Debt $\rightarrow$ unsafe or unethical autonomous decisions
    \item Documentation Debt $\rightarrow$ lack of documentation of agent behavior
\end{enumerate}

\begin{table*}[t]
\centering
\caption{Direct Transformation of AITDs into AgTDs}
\label{tab:direct_transformation}
\scriptsize
\begin{tabular}{p{1.6cm} p{3.5cm} p{5.5cm} p{3.5cm}}
\toprule
\textbf{AITD} & \textbf{Original Definition} & \textbf{Transformation Mechanism in Agentic AI} & \textbf{Resulting AgTD (Agentic Manifestation)} \\
\midrule

Data Debt &
Poor-quality, incomplete, inconsistent, or outdated data used for training and operation &
Shared memory stores, vector databases, and knowledge repositories propagate faulty information across multiple agents. &
Corrupted or incomplete shared agent memory \\

Pipeline Jungle &
Overly complex, poorly structured, and difficult-to-maintain data pipelines &
Multi-agent workflows introduce chained task execution, tool invocations, and inter-agent dependencies that increase pipeline complexity. &
Complex multi-agent data pipelines \\

Dispensable Dependency &
Unnecessary or redundant software libraries, frameworks, or external dependencies &
Heavy reliance on external tools, APIs, and services amplifies dependency management challenges and operational overhead. &
Unnecessary tool/API reliance \\

Scattered ML Libraries &
Use of multiple heterogeneous and incompatible machine learning libraries &
Agent ecosystems often combine diverse frameworks, models, and orchestration platforms, creating interoperability challenges. &
Fragmented agent frameworks \\

Glue Code &
Ad hoc integration logic connecting heterogeneous software components &
Extensive coordination among agents, tools, memory stores, and services increases orchestration complexity. &
Fragile agent orchestration logic \\

Debug Code (UDC) &
Temporary debugging artifacts unintentionally retained in production systems &
Prompts, traces, logs, and testing instructions may remain embedded within deployed agents and workflows. &
Leftover debug prompts/logs in agents \\

Ethical Debt &
Deferred consideration of fairness, transparency, accountability, and ethical concerns &
Autonomous reasoning and decision-making amplify unresolved ethical issues during runtime operation. &
Unsafe or unethical autonomous decisions \\

Documentation Debt &
Missing, incomplete, outdated, or inconsistent documentation &
Agent reasoning processes, memory interactions, and tool orchestration become difficult to understand and maintain without adequate documentation. &
Lack of documentation of agent behavior \\

\bottomrule
\end{tabular}
\end{table*}

\subsubsection{Contextual Transformation (c)}
In contextual transformation, the technical debt retains its core nature and underlying cause, but its manifestation evolves due to the unique characteristics of Agentic AI systems, particularly autonomy, interaction, coordination, and adaptive decision-making. Unlike direct transformation, where the debt is largely preserved in its original form, contextual transformation reflects a shift in how the debt is expressed and experienced within dynamic, multi-agent environments. This semantic arises when traditional technical debt interacts with agent-specific properties such as decentralized control, continuous feedback, and inter-agent communication. As agents operate autonomously and collaborate to achieve shared or individual goals, previously localized issues become embedded within interaction patterns, coordination mechanisms, and decision workflows. Consequently, the same underlying debt may lead to qualitatively different behaviors, often increasing in complexity and impact without fundamentally altering its conceptual origin.

Contextual transformations are particularly evident in debts related to system behavior, coordination, and interaction logic, such as feedback loops, implicit dependencies, or duplicated functionality. In agentic settings, these debts no longer manifest as isolated inefficiencies but instead influence how agents communicate, adapt, and respond to one another. For instance, a hidden feedback loop in a traditional system may evolve into a self-reinforcing interaction cycle among agents, while undeclared dependencies may result in implicit coordination failures across distributed workflows.

To improve traceability and clarify the rationale behind each transformation, Table~\ref{tab:contextual_transformation} summarizes the original AITDs, their definitions, the mechanisms through which they are recontextualized in Agentic AI systems, and their resulting AgTD manifestations. Unlike direct transformations, these debts retain their underlying causes but evolve in how they are expressed due to agent autonomy, coordination, reasoning, and interaction dynamics.

This category highlights that the transition to Agentic AI does not merely scale existing technical debt but recontextualizes it within a more complex operational environment. As a result, contextual transformation emphasizes the need to analyze technical debt not only in terms of its structural origin but also in relation to its behavioral and interaction-driven manifestations, which are central to the functioning of agentic systems.

\textbf{AgTDs under Contextual Transformation:}
\begin{enumerate}
    \item Duplicate Model Code $\rightarrow$ duplicated capabilities across agents
    \item Hidden Feedback Loops $\rightarrow$ self-reinforcing agent loops
    \item Undeclared Consumers $\rightarrow$ implicit agent-to-agent dependencies
    \item Algorithm Bias $\rightarrow$ biased reasoning in agent decisions
    \item AI Architectural Debt $\rightarrow$ poorly designed multi-agent architecture
    \item Jumbled Model Architecture $\rightarrow$ disorganized agent roles/structures
    \item Configuration Debt $\rightarrow$ misconfigured agent parameters/tools
    \item Cultural/People Debt $\rightarrow$ poor human-agent collaboration
    \item Test Debt $\rightarrow$ insufficient testing of agent interactions and behaviors
\end{enumerate}

\begin{table*}[t]
\centering
\caption{Contextual Transformation of AITDs into  AgTDs}
\label{tab:contextual_transformation}
\scriptsize
\begin{tabular}{p{1.6cm} p{3.5cm} p{5.5cm} p{3.5cm}}
\toprule
\textbf{AITD} & \textbf{Original Definition} & \textbf{Transformation Mechanism in Agentic AI} & \textbf{Resulting AgTD (Agentic Manifestation)} \\
\midrule

Duplicate Model Code &
Redundant implementations of similar model functionality across different components &
Agentic systems often deploy multiple specialized agents that independently implement similar reasoning, planning, or execution capabilities, leading to functional duplication and maintenance overhead. &
Duplicated capabilities across agents \\

Hidden Feedback Loops &
Undetected circular dependencies that reinforce system behavior and obscure causal relationships &
Agents continuously exchange information and react to one another’s outputs, creating recursive interaction cycles that amplify behaviors and propagate errors. &
Self-reinforcing agent loops \\

Undeclared Consumers &
Dependencies created by downstream consumers that are not explicitly documented or managed &
Agents frequently rely on outputs, memory states, or services produced by other agents without explicit dependency specifications, reducing transparency and maintainability. &
Implicit agent-to-agent dependencies \\

Algorithm Bias &
Biases embedded within algorithms, models, or decision-making processes &
Biased reasoning patterns can propagate through autonomous planning and decision-making processes, influencing agent actions and recommendations. &
Biased reasoning in agent decisions \\

AI Architectural Debt &
Suboptimal architectural decisions that hinder scalability, maintainability, or evolution of AI systems &
The introduction of multiple interacting agents, shared memory, and tool orchestration increases architectural complexity and magnifies the impact of poor design choices. &
Poorly designed multi-agent architecture \\

Jumbled Model Architecture &
Disorganized model structures, unclear responsibilities, or poorly modularized designs &
Agentic systems require clear role separation and coordination mechanisms. Poor organization of agent responsibilities leads to inefficient communication and execution. &
Disorganized agent roles and structures \\

Configuration Debt &
Improperly managed, inconsistent, or outdated system configurations &
Agentic systems require coordinated management of prompts, memory settings, tool permissions, reasoning policies, and execution parameters across multiple agents. &
Misconfigured agent parameters and tools \\

Cultural/People Debt &
Organizational, communication, or collaboration issues among development teams and stakeholders &
The integration of autonomous agents into human workflows introduces new collaboration challenges, including unclear responsibilities, trust issues, and ineffective oversight. &
Poor human-agent collaboration \\

Test Debt &
Insufficient, outdated, or incomplete testing practices &
Agent interactions, emergent behaviors, adaptive reasoning, and dynamic tool usage create testing challenges that extend beyond traditional unit and integration testing. &
Insufficient testing of agent interactions and behaviors \\

\bottomrule
\end{tabular}
\end{table*}

\subsubsection{Manifestation Expansion (e)}

In manifestation expansion, the technical debt undergoes a fundamental transformation, evolving beyond its original form into broader, system-level, and emergent behaviors that were not present in traditional or even conventional AI systems. Unlike direct and contextual transformations, where the core nature of the debt is preserved, manifestation expansion reflects a qualitative shift in which the debt becomes intrinsically tied to the collective behavior of the agentic system as a whole. This semantic arises when the interaction of autonomy, distributed execution, continuous adaptation, and multi-agent coordination amplifies existing technical debt into new forms of systemic risk. In such cases, the original debt is no longer confined to a specific component, interaction, or context; instead, it manifests as a property of the entire system, often emerging from complex dependencies, feedback mechanisms, and dynamic decision processes. As a result, the effects of the debt become non-linear, difficult to predict, and increasingly challenging to localize or isolate.

Manifestation expansion is particularly evident in debts related to architecture, system integration, and evaluation, where the combination of multiple agents, tools, and evolving states gives rise to behaviors such as cascading failures, tightly coupled dependency networks, responsibility leakage, and unstable system-wide dynamics. For example, architectural inconsistencies may evolve into large-scale coordination breakdowns, while compatibility issues can propagate into persistent integration failures across agent ecosystems. Similarly, weak evaluation strategies may expand into an inability to reliably assess system performance under diverse and evolving conditions.

To improve traceability and clarify the rationale behind each transformation, Table~\ref{tab:manifestation_expansion} summarizes the original AITDs, their definitions, the mechanisms through which they expand within Agentic AI systems, and their resulting AgTD manifestations. Unlike direct and contextual transformations, manifestation expansion occurs when existing technical debts evolve into broader system-level behaviors and emergent risks driven by autonomy, adaptation, multi-agent coordination, and runtime interactions.

This category underscores the transition from technical debt as a design or implementation concern to a \textit{systemic and emergent phenomenon}, where the debt is embedded within the collective intelligence and behavior of the system. Consequently, manifestation expansion highlights the limitations of traditional technical debt management approaches, which are typically localized and reactive, and emphasizes the need for holistic, system-aware strategies that account for emergence, propagation, and long-term evolution in Agentic AI environments.

\textbf{AgTDs under Manifestation Expansion:}
\begin{enumerate}
    \item Dead Experimental Code $\rightarrow$ legacy or unused agents/tools
    \item Correction Cascades $\rightarrow$ cascading multi-agent failures
    \item Entanglement $\rightarrow$ tightly coupled multi-agent dependencies
    \item Deep God File $\rightarrow$ monolithic “super-agent” systems
    \item Overly Simplified Metrics $\rightarrow$ weak evaluation of agent behavior
    \item Compatibility Debt $\rightarrow$ incompatible tools/models across agents
    \item Abstraction Debt $\rightarrow$ poor abstraction of agent interfaces/APIs
    \item Boundary Erosion $\rightarrow$ responsibility leakage across agents
    \item Requirement Debt $\rightarrow$ unclear or misaligned agent goals
    \item Defect Debt $\rightarrow$ runtime failures (hallucinations, tool misuse)
    \item Versioning Debt $\rightarrow$ difficulty tracking prompts/tools/versions
    \item Process/Infrastructure Debt $\rightarrow$ inadequate agent deployment infrastructure
    \item Build Debt $\rightarrow$ inconsistent agent system builds
    \item MLS (Multiple Language Smell) $\rightarrow$ inconsistent representations across agent languages or modalities

\end{enumerate}

\begin{table*}[t]
\centering
\caption{Manifestation Expansion of AITDs into AgTDs}
\label{tab:manifestation_expansion}
\scriptsize
\begin{tabular}{p{1.6cm} p{3.5cm} p{5.5cm} p{3.5cm}}
\toprule
\textbf{AITD} & \textbf{Original Definition} & \textbf{Transformation Mechanism in Agentic AI} & \textbf{Resulting AgTD (Agentic Manifestation)} \\
\midrule

Dead Experimental Code &
Obsolete, unused, or abandoned code artifacts retained in production systems &
Agentic systems often accumulate dormant agents, unused tools, deprecated workflows, and inactive reasoning modules that remain integrated despite no longer serving operational purposes. &
Legacy or unused agents/tools \\

Correction Cascades &
A defect fix in one component triggers unintended issues in other dependent components &
Autonomous agents continuously exchange outputs and decisions, allowing errors or corrective actions in one agent to propagate across interconnected workflows. &
Cascading multi-agent failures \\

Entanglement &
Strong and implicit dependencies among components that make changes difficult and risky &
Shared memory, tool dependencies, and coordinated workflows create tightly interconnected agent ecosystems with limited modularity. &
Tightly coupled multi-agent dependencies \\

Deep God File &
An excessively large and centralized component responsible for multiple unrelated functions &
Organizations may consolidate planning, reasoning, orchestration, and execution into a single dominant agent, creating bottlenecks and maintainability challenges. &
Monolithic ``super-agent'' systems \\

Overly Simplified Metrics &
Evaluation metrics that fail to capture system complexity and real-world performance &
Traditional metrics focus on model accuracy or task completion while overlooking coordination quality, reasoning effectiveness, trustworthiness, and emergent behaviors. &
Weak evaluation of agent behavior \\

Compatibility Debt &
Difficulties arising from incompatible technologies, frameworks, versions, or interfaces &
Agentic ecosystems integrate diverse models, tools, APIs, and frameworks whose incompatibilities disrupt collaboration and execution. &
Incompatible tools/models across agents \\

Abstraction Debt &
Poorly designed abstractions that obscure responsibilities and increase complexity &
Weakly defined interfaces between agents, tools, memory systems, and services create ambiguity and hinder scalability. &
Poor abstraction of agent interfaces/APIs \\

Boundary Erosion &
Gradual loss of clear separation between system responsibilities and ownership boundaries &
Agents dynamically share tasks, memory, and decisions, making it difficult to distinguish ownership, accountability, and operational responsibilities. &
Responsibility leakage across agents \\

Requirement Debt &
Incomplete, ambiguous, outdated, or poorly specified requirements &
High-level objectives provided to agents may be vague, conflicting, or misaligned, leading to inconsistent planning and execution. &
Unclear or misaligned agent goals \\

Defect Debt &
Accumulation of known defects that remain unresolved &
Autonomous reasoning, tool usage, and adaptive execution introduce runtime failures that may persist and propagate if not systematically addressed. &
Runtime failures (hallucinations, tool misuse) \\

Versioning Debt &
Poor management of software, model, configuration, or dependency versions &
Agentic systems continuously evolve through prompt updates, tool changes, model upgrades, and memory modifications, complicating traceability and reproducibility. &
Difficulty tracking prompts, tools, and versions \\

Process/Infrastructure Debt &
Suboptimal operational processes and infrastructure that hinder system evolution and maintenance &
Agent deployment requires orchestration platforms, monitoring frameworks, memory services, and governance mechanisms that may not scale with system complexity. &
Inadequate agent deployment infrastructure \\

Build Debt &
Inefficient, inconsistent, or unreliable build and deployment processes &
Distributed agent ecosystems require coordinated deployment of agents, tools, prompts, memory stores, and supporting services across environments. &
Inconsistent agent system builds \\

Multiple Language Smell &
Complexity arising from the use of multiple programming languages, representations, or technologies &
Agents may communicate through different languages, modalities, schemas, embeddings, or reasoning representations, creating interoperability and consistency challenges. &
Inconsistent representations across agent languages or modalities \\

\bottomrule
\end{tabular}
\end{table*}

\noindent These transformation semantics collectively demonstrate a clear evolution from traditional, component-level technical debt toward dynamic, distributed, and emergent system-level risks in Agentic AI systems. In particular, they reveal that technical debt is no longer confined to static artifacts such as code modules, data pipelines, or architectural components, but instead becomes embedded within the interactions, coordination mechanisms, and adaptive behaviors of autonomous agents operating in complex environments. As systems transition from single-model pipelines to multi-agent ecosystems, technical debt progressively shifts from being localized and predictable to being interaction-driven, continuously evolving, and often non-linear in its manifestation. Moreover, the distinction between direct, contextual, and manifestation expansion semantics provides a structured lens for understanding how different categories of debt evolve under agentic conditions. While some debts are inherited with minimal change, others are reshaped by the dynamics of agent interaction, and a significant portion expands into higher-order phenomena that emerge only at the system level. This progression highlights the increasing difficulty of isolating, diagnosing, and mitigating technical debt, as its effects become intertwined with system behavior and collective intelligence rather than individual components. 
Ultimately, these semantics provide a structured framework for understanding how technical debt evolves in Agentic AI systems and why its identification, diagnosis, and mitigation become increasingly challenging in autonomous, multi-agent environments. This insight motivates the need for new analytical and management approaches tailored to the complexity and emergent behaviors of agentic systems.


\subsection{Categorization and Analysis of AgTDs by Root-Cause Taxonomy}

To provide a structured understanding of Agentic Technical Debt, we analyze all identified AgTDs across the seven root-cause categories inherited from the original AITD taxonomy. This categorization ensures conceptual consistency while enabling a systematic examination of how technical debt manifests and evolves in agentic systems. For each AgTD, we provide its definition, impact, use case, and example, offering both theoretical clarity and practical relevance. 
Before presenting the individual AgTD categories, it is important to clarify that the debts identified in this study are not defined by the resulting failures, risks, vulnerabilities, or system complexities themselves. Rather, they represent deferred, suboptimal, or insufficiently addressed design, implementation, architectural, coordination, or governance decisions whose consequences manifest as increased maintenance, evolution, operational, monitoring, or remediation costs in Agentic AI systems. Accordingly, phenomena such as cascading failures, coordination breakdowns, or inconsistent agent behavior are viewed as manifestations or consequences of underlying technical debt rather than the debt itself. This taxonomy-driven analysis supports the identification of category-specific risks and informs potential mitigation strategies in Agentic AI environments.

\subsubsection{Data and Library-Related Debts}

Debts arising from poor data quality, fragmented data pipelines, and inefficient or excessive dependency management in agentic systems, particularly affecting shared memory, knowledge exchange, and data-driven decision-making. In Agentic AI environments, where multiple agents continuously access and update shared data sources, these debts become amplified and propagate across agents, influencing system-wide behavior. These debts are critical because data underpins agent perception, reasoning, and action. Inconsistencies, redundancy, or inefficiencies in data handling can lead to misaligned behaviors, conflicting outputs, and degraded performance. Additionally, the use of heterogeneous libraries and external dependencies introduces challenges in maintaining consistency and reliability. In practice, such debts arise in multi-agent systems relying on shared vector databases, APIs, or distributed pipelines. For example, inconsistent embeddings in a shared knowledge store may cause agents to produce conflicting responses, while excessive dependency on external tools can increase latency and reduce system robustness.

\renewcommand{\labelenumi}{\roman{enumi}.}

\begin{enumerate}

\item \textit{Data Debt:} Refers to inconsistencies, incompleteness, redundancy, or corruption in shared data sources and knowledge representations used by agents, including training data, embeddings, and memory stores. In agentic systems, this debt is further amplified due to continuous data updates and shared access across multiple agents.  
\textit{Impact:} Leads to conflicting reasoning, degraded decision accuracy, propagation of errors across agents, and unreliable system outputs.  
\textit{Use Case:} Multi-agent systems sharing vector databases, knowledge graphs, or external data repositories.  
\textit{Example:} Agents retrieving outdated or misaligned embeddings from a shared memory store, resulting in contradictory answers or inconsistent reasoning paths.

\par \textit{Source(s):} \cite{sculley2015hidden, alahdab2019empirical, zhang2022code, tang2021empirical, arpteg2018software, polyzotis2018data, foidl2022data, lenarduzzi2021software, foidl2019technical, breck2017ml, hutchinson2021towards, moreschini2024towards, nahar2022collaboration, shivashankar2022maintainability, roselli2019managing, cote2024quality, belani2019requirements, sas2023architectural, wang2023technical, khanvilkar2025automated, recupito2024unmasking, cunha2020investigating, de2025software, akgul2025aligning, ximenes2025investigating, shome2022data, moldovan2024python}.

\item \textit{Pipeline Jungle:} Represents overly complex, deeply nested, and poorly structured data and control pipelines across multiple agents, often involving chained tools, APIs, and intermediate transformations. Such complexity arises from incremental system evolution without adequate refactoring.  
\textit{Impact:} Reduces observability, increases debugging difficulty, slows fault diagnosis, and obscures the root causes of failures across agent workflows.  
\textit{Use Case:} Systems integrating multiple agents with sequential or parallel tool invocations and data transformations.  
\textit{Example:} A failure in an upstream agent silently propagates through several downstream agents due to lack of pipeline transparency and monitoring.

\par \textit{Source(s):} \cite{recupito2024technical, sculley2015hidden, alahdab2019empirical, washizaki2019studying, foidl2022data, moreschini2024towards, shivashankar2022maintainability, belani2019requirements, wang2023technical}.

\item \textit{Dispensable Dependency:} Refers to unnecessary or redundant reliance on external tools, APIs, libraries, or services that do not provide significant added value to agent functionality. These dependencies often accumulate over time due to rapid development and integration practices.  
\textit{Impact:} Increases system latency, operational cost, maintenance overhead, and susceptibility to external failures or changes.  
\textit{Use Case:} Agents invoking multiple external services for similar or overlapping tasks.  
\textit{Example:} An agent making repeated API calls to different services for identical data retrieval, resulting in inefficiency and increased response time.

\par \textit{Source(s):} \cite{chaudhary2018review, 10628360, breck2017ml}.

\item \textit{Scattered ML Libraries:} Occurs when agents rely on heterogeneous, fragmented, or incompatible machine learning libraries and frameworks, often due to independent development or lack of standardization.  
\textit{Impact:} Leads to inconsistent outputs, integration challenges, increased maintenance effort, and reduced system coherence.  
\textit{Use Case:} Multi-agent systems where different agents are built using diverse ML frameworks or toolchains.  
\textit{Example:} Agents producing varying predictions or behaviors due to differences in underlying libraries, model implementations, or runtime environments.

\par \textit{Source(s):} \cite{recupito2024technical, 10628360}.

\end{enumerate}

\subsubsection{Model and Code-Related Debts}

Debts related to implementation complexity, coupling, maintainability, and code-level design in agentic systems. In Agentic AI environments, these debts are significantly amplified by the distributed, autonomous, and interaction-driven nature of agents, where code-level inefficiencies and design flaws not only affect individual components but also propagate across agents, influencing coordination, execution flows, and overall system stability.

\begin{enumerate}

\item \textit{Glue Code:} Refers to fragile, often ad hoc orchestration logic that connects agents, tools, and execution workflows, typically developed without standardized interfaces or abstraction layers. In agentic systems, such glue code becomes a critical component governing coordination and interaction.  
\textit{Impact:} Leads to brittle integrations, reduced system robustness, and frequent execution failures when underlying tools, APIs, or interfaces evolve.  
\textit{Use Case:} Tool-integrated LLM agents coordinating multiple services and APIs.  
\textit{Example:} A minor change in an external API causing the breakdown of an entire multi-agent workflow.

\par \textit{Source(s):}  \cite{recupito2024technical, sculley2015hidden, alahdab2019empirical, 10628360, tang2021empirical, washizaki2019studying, van2021prevalence, arpteg2018software, li2023debtviz, li2023automatic, lenarduzzi2021software, obrien202223, perez2021technical, moreschini2024towards, shivashankar2022maintainability, belani2019requirements, bavota2016large, wang2023technical}. 

\item \textit{Duplicate Model Code:} Represents redundant implementations of similar logic or functionality across multiple agents due to lack of modularization or reuse mechanisms.  
\textit{Impact:} Increases maintenance effort, introduces inconsistencies across agents, and complicates system updates and evolution.  
\textit{Use Case:} Multi-agent systems performing similar reasoning or processing tasks independently.  
\textit{Example:} Multiple agents implementing identical summarization or classification logic without shared modules.

\par \textit{Source(s):}  \cite{tang2021empirical, albuquerque2022comprehending, van2021prevalence, li2023debtviz, li2023automatic, obrien202223, perez2021technical, jebnoun2022clones, khanvilkar2025automated, li2022identifying}. 

\item \textit{Dead Experimental Code:} Refers to obsolete, unused, or partially integrated agents, tools, or experimental components that persist in the system after development or testing phases.  
\textit{Impact:} Adds unnecessary complexity, increases resource consumption, and reduces system clarity and maintainability.  
\textit{Use Case:} Rapid prototyping and iterative development environments involving frequent experimentation.  
\textit{Example:} Deprecated agents or experimental modules remaining in the production pipeline and occasionally being triggered.

\par \textit{Source(s):} \cite{sculley2015hidden, tang2021empirical, washizaki2019studying, li2023debtviz, li2023automatic, obrien202223, perez2021technical, moreschini2024towards, belani2019requirements, wang2023technical, li2022identifying}.


\item \textit{Hidden Feedback Loops:} Arises when feedback mechanisms among agents are insufficiently designed, monitored, or governed, resulting in recursive interactions that remain undetected or unmanaged.
\textit{Impact:} Amplifies errors, biases, and instability over time, increasing the future cost of debugging, maintenance, and system governance. \textit{Use Case:} Iterative reasoning, retrieval-augmented, or self-refinement agent systems. \textit{Example:} Feedback loops created by agents repeatedly validating each other's outputs without adequate monitoring or corrective mechanisms, leading to the accumulation of future remediation costs.

\par \textit{Source(s)}  \cite{sculley2015hidden, arpteg2018software, menshawy2024navigating, moreschini2024towards, shivashankar2022maintainability, roselli2019managing, khritankov2021hidden, wang2023technical, shukla2022challenges}.

\item \textit{Correction Cascades:} Arises when dependencies among agents are insufficiently designed, validated, or isolated, causing errors or corrective actions in one agent to propagate through downstream agents. 
\textit{Impact:} Increases future maintenance and remediation costs by amplifying failures across agent workflows and reducing system reliability.
\textit{Use Case:} Sequential or pipeline-based multi-agent decision systems.
\textit{Example:} An inadequately validated intermediate output propagating through multiple agents, requiring costly system-wide corrections.
\par  \textit{Source(s):} \cite{recupito2024technical, sculley2015hidden, belani2019requirements, shukla2022challenges}.

\item \textit{Entanglement:} Arises from architectural decisions that create excessive coupling and implicit dependencies among agents, shared memory, or execution contexts.
\textit{Impact:} Reduces modularity, flexibility, and system resilience, making future modifications, debugging, and maintenance more costly.
\textit{Use Case:} Highly interconnected multi-agent ecosystems with shared execution contexts.
\textit{Example:} An agent designed with strong dependencies on multiple other agents, causing widespread failures when a single component changes.
\par \textit{Source(s):}  \cite{sculley2015hidden, menshawy2024navigating, belani2019requirements, wang2023technical}.

\item \textit{Undeclared Consumers:} Occurs when dependencies between agents are insufficiently documented, specified, or governed, allowing agents to consume outputs without explicit interfaces or contracts.
\textit{Impact:} Increases integration, maintenance, and debugging effort due to hidden dependencies and unpredictable interactions.
\textit{Use Case:} Systems with loosely coupled agents sharing intermediate outputs.
\textit{Example:} An agent relying on another agent’s output format without a formally defined and maintained interface specification.
\par \textit{Source(s):} \cite{recupito2024technical, sculley2015hidden, chaudhary2018review, washizaki2019studying, belani2019requirements, wang2023technical}.

\item \textit{Deep God File:} Arises from architectural decisions that concentrate excessive responsibilities within a single agent rather than distributing functionality appropriately across the system.
\textit{Impact:} Limits scalability, increases maintenance complexity, and creates a costly single point of failure.
\textit{Use Case:} Centralized orchestration architectures with overloaded agents.
\textit{Example:} A single agent responsible for planning, reasoning, coordination, and execution due to deferred architectural refactoring.
\par  \textit{Source(s):}  \cite{recupito2024technical, cunha2020investigating}.

\item \textit{Debug Code (UDC):} Occurs when temporary debugging logic, prompts, traces, or code paths are intentionally retained or insufficiently removed during deployment.
\textit{Impact:} Increases maintenance overhead, introduces operational noise, and may expose unintended information or behaviors.
\textit{Use Case:} Development-to-production transitions in agent-based systems.
\textit{Example:} Debug prompts or logging instructions remaining active in deployed agents due to deferred cleanup activities.

\textit{Source(s):} \cite{recupito2024technical, li2023debtviz, li2023automatic, obrien202223, perez2021technical}.

\item \textit{Multiple Language Smell (MLS):} Arises when heterogeneous representations, languages, prompts, APIs, or interaction modalities are introduced without sufficient standardization, coordination, or governance.
\textit{Impact:} Increases maintenance effort and integration complexity by creating inconsistencies in communication and reasoning across agent components.
\textit{Use Case:} Multi-modal or multi-interface agent systems combining different interaction formats.
\textit{Example:} Conflicting behaviors resulting from inconsistencies between natural-language prompts, code-based instructions, and API specifications that were not adequately harmonized during system design.

\textit{Source(s):}  \cite{recupito2024technical, sculley2015hidden, alahdab2019empirical, tang2021empirical, washizaki2019studying, moreschini2024towards, wang2023technical}.

\end{enumerate}

\subsubsection{Algorithm-Related Debts}

Debts arising from model behavior, bias, and limitations in evaluation strategies. In Agentic AI systems, these debts are particularly critical as they directly influence how agents perceive, reason, and make decisions in autonomous and interactive environments. Unlike traditional systems, where algorithmic issues may remain confined to model outputs, in agentic settings such debts propagate through agent interactions and can significantly impact system-wide behavior and outcomes.

\begin{enumerate}

\item \textit{Algorithm Bias:} Refers to systematic bias embedded in agent reasoning, decision-making processes, or learned representations, often originating from skewed training data or model design choices.  
\textit{Impact:} Produces unfair, discriminatory, or harmful outcomes, potentially affecting user trust and system reliability.  
\textit{Use Case:} Decision-support systems in domains such as healthcare, finance, or recruitment.  
\textit{Example:} Agents recommending biased treatment plans or hiring decisions due to imbalanced training data.

\par \textit{Source(s):} \cite{chaudhary2018review, liu2020using, chen2023toward, arpteg2018software, liu2021exploratory, simon2023algorithm, wang2023technical, nikanjam2021design, li2022identifying, de2025software}.

\item \textit{Ethical Debt:} Represents the accumulation of risks associated with unethical, unsafe, or misaligned agent behaviors that are not adequately addressed during design and deployment.  
\textit{Impact:} Leads to regulatory, legal, and societal risks, as well as reduced user trust and acceptance.  
\textit{Use Case:} Autonomous assistants, decision-making agents, and human-facing AI systems.  
\textit{Example:} Agents generating harmful, misleading, or inappropriate advice in sensitive domains.

\par \textit{Source(s):} \cite{menshawy2024navigating, chang2022understanding, roselli2019managing, petrozzino2021pays}.

\item \textit{Overly Simplified Metrics:} Refers to the use of inadequate or narrow evaluation metrics that fail to capture the complexity of agent behaviors and system objectives.  
\textit{Impact:} Misguides optimization, leading to suboptimal or unintended system behaviors and poor real-world performance.  
\textit{Use Case:} Reward-based learning systems and performance evaluation pipelines.  
\textit{Example:} Agents optimized for simple accuracy metrics while neglecting fairness, robustness, or long-term impact.

\par \textit{Source(s):}~\cite{chaudhary2018review, khanvilkar2025automated, ximenes2025investigating}.

\end{enumerate}

\subsubsection{Design and Architecture-Related Debts}

Debts related to system architecture, modularity, component design, and agent boundaries. In Agentic AI systems, these debts are particularly critical as they shape how agents are organized, coordinated, and integrated within the system. Poor architectural decisions can lead to inefficiencies not only at the component level but also across agent interactions, resulting in reduced scalability, coordination breakdowns, and increased system fragility.

\begin{enumerate}

\item \textit{AI Architectural Debt:} Refers to poorly designed multi-agent system structures, including inefficient communication patterns, lack of hierarchy, or inadequate modularization.  
\textit{Impact:} Limits scalability, reduces coordination efficiency, and increases system complexity as the number of agents grows.  
\textit{Use Case:} Distributed agent architectures requiring structured communication and task allocation.  
\textit{Example:} Inefficient communication hierarchies causing delays and redundant message passing among agents.

\par \textit{Source(s):}  \cite{liu2020using, albuquerque2022comprehending, li2023debtviz, li2023automatic, perez2021technical, jebnoun2022clones, bavota2016large, yan2018automating, liu2021exploratory, nikanjam2021design, li2022identifying, de2025software, sutoyo2024satdaug}.

\item \textit{Jumbled Model Architecture:} Represents disorganized agent roles, unclear responsibilities, and poorly structured interactions among agents.  
\textit{Impact:} Reduces system clarity, increases maintenance difficulty, and leads to inefficient execution flows.  
\textit{Use Case:} Complex multi-agent workflows with overlapping or poorly defined roles.  
\textit{Example:} Multiple agents performing similar or conflicting tasks due to lack of clear role definition.

\par \textit{Source(s):}  \cite{recupito2024technical, albuquerque2022comprehending, li2023automatic, foidl2019technical, perez2021technical, sas2023architectural, li2022identifying, perez2019proposed, cunha2020investigating}. 

\item \textit{Compatibility Debt:} Refers to incompatibilities across tools, models, communication protocols, or APIs used by different agents.  
\textit{Impact:} Causes integration failures, disrupts communication, and limits interoperability between system components.  
\textit{Use Case:} Multi-tool agent ecosystems combining heterogeneous technologies.  
\textit{Example:} Agents failing to exchange information due to mismatched API formats or data schemas.

\par \textit{Source(s):}  \cite{liu2020using, chen2023toward, lenarduzzi2021software, liu2021exploratory}.

\item \textit{Abstraction Debt:} Occurs when agent interfaces and interaction mechanisms lack proper abstraction, leading to tightly coupled implementations.  
\textit{Impact:} Reduces flexibility, reusability, and extensibility of the system.  
\textit{Use Case:} API-driven or modular agent systems requiring clear separation of concerns.  
\textit{Example:} Hard-coded communication logic between agents that prevents reuse or adaptation.

\par \textit{Source(s):}  \cite{sculley2015hidden, 10628360, tang2021empirical, washizaki2019studying}.

\item \textit{Boundary Erosion:} Refers to the blurring or weakening of boundaries between agents, resulting in overlapping responsibilities and unclear ownership.  
\textit{Impact:} Leads to coordination breakdown, role confusion, and reduced system reliability.  
\textit{Use Case:} Multi-agent collaboration systems with shared tasks and responsibilities.  
\textit{Example:} Agents unintentionally performing tasks outside their intended roles, causing inconsistencies.

\par \textit{Source(s):}  \cite{sculley2015hidden, chaudhary2018review, tang2021empirical}.

\item \textit{Configuration Debt:} Refers to misconfigured parameters, tools, or environment settings that affect agent behavior and interactions.  
\textit{Impact:} Causes unstable system behavior, inconsistent outputs, and reduced performance.  
\textit{Use Case:} Prompt-driven or parameterized agent systems requiring fine-tuned configurations.  
\textit{Example:} Incorrect parameter settings leading to unexpected agent responses or degraded reasoning quality.

\par \textit{Source(s):}  \cite{sculley2015hidden, alahdab2019empirical, zhang2022code, tang2021empirical, chen2023toward, foidl2019technical, breck2017ml, jebnoun2022clones, belani2019requirements, khanvilkar2025automated, nikanjam2021design, ximenes2025investigating}.

\end{enumerate}

\subsubsection{Operational and Lifecycle-Related Debts}

Debts related to deployment, configuration, versioning, monitoring, and overall system operations throughout the lifecycle of agentic systems. In Agentic AI environments, these debts are particularly critical due to the continuous evolution of agents, dynamic interaction patterns, and reliance on external tools and infrastructure. Unlike traditional systems, operational issues in agentic settings can propagate rapidly across agents, affecting system stability, reproducibility, and long-term maintainability.

\begin{enumerate}

\item \textit{Requirement Debt:} Refers to poorly defined, incomplete, or ambiguous agent goals, tasks, or constraints, often resulting from insufficient requirement specification or evolving system objectives.  
\textit{Impact:} Leads to misaligned outputs, inefficient agent behavior, and inability to meet intended system objectives.  
\textit{Use Case:} Goal-driven agent systems where agents must autonomously interpret and execute tasks.  
\textit{Example:} Agents solving incorrect or irrelevant tasks due to vague or misinterpreted instructions.

\par \textit{Source(s):}  \cite{liu2020using, albuquerque2022comprehending, li2023debtviz, li2023automatic, obrien202223, perez2021technical, moreschini2024towards, nahar2022collaboration, belani2019requirements, bavota2016large, liu2021exploratory, li2022identifying, de2025software, sutoyo2024satdaug}.

\item \textit{Defect Debt:} Refers to runtime issues such as hallucinations, incorrect reasoning, or misuse of tools that arise during agent execution.  
\textit{Impact:} Produces incorrect, inconsistent, or unsafe outputs, potentially affecting downstream agents and overall system reliability.  
\textit{Use Case:} LLM-based agents performing reasoning, planning, or tool invocation tasks.  
\textit{Example:} Agents generating hallucinated responses or incorrectly invoking tools, leading to faulty outputs.

\par \textit{Source(s):}  \cite{liu2020using, albuquerque2022comprehending, li2023automatic, obrien202223, perez2021technical, bavota2016large, yan2018automating, liu2021exploratory, li2022identifying}.

\item \textit{Versioning Debt:} Refers to the difficulty in managing evolving versions of prompts, models, tools, and configurations across agents and environments.  
\textit{Impact:} Reduces reproducibility, traceability, and consistency of system behavior over time.  
\textit{Use Case:} Continuous integration and deployment pipelines for agent-based systems.  
\textit{Example:} Different agents using inconsistent prompt or model versions, leading to varying outputs across environments.

\par \textit{Source(s):} \cite{washizaki2019studying, albuquerque2022comprehending, perez2021technical, cote2024quality, shukla2022challenges}. 

\item \textit{Process/Infrastructure Debt:} Refers to weaknesses in the operational infrastructure, including orchestration frameworks, deployment pipelines, and monitoring mechanisms supporting agent execution.  
\textit{Impact:} Limits scalability, reduces system reliability, and hinders efficient operation and maintenance.  
\textit{Use Case:} Production-level deployment of multi-agent systems requiring robust orchestration and monitoring.  
\textit{Example:} Poorly designed orchestration pipelines leading to inefficient task scheduling and execution failures.

\textit{Source(s):} \cite{lenarduzzi2021software, perez2021technical, nahar2022collaboration}.

\item \textit{Build Debt:} Refers to inconsistencies or inefficiencies in system build processes across different environments, often due to lack of standardization or automation.  
\textit{Impact:} Causes deployment instability, non-reproducible results, and increased debugging effort.  
\textit{Use Case:} CI/CD pipelines for agent-based systems deployed across multiple environments.  
\textit{Example:} Agents producing different outputs in development and production environments due to inconsistent build configurations.

\par \textit{Source(s):}  \cite{li2023automatic, perez2021technical, li2022identifying}.

\end{enumerate}

\subsubsection{Documentation and Communication Debts} 

Debts arising from insufficient documentation and ineffective communication between human stakeholders and agentic systems. In Agentic AI environments, where agents operate autonomously and interact dynamically with users and other agents, these debts become particularly critical. A lack of clear documentation and communication mechanisms not only hinders system understanding but also reduces transparency, explainability, and user trust, especially in complex multi-agent settings.

\begin{enumerate}

\item \textit{Documentation Debt:} Refers to missing, incomplete, or outdated documentation describing agent behavior, reasoning processes, system workflows, and interaction protocols.  
\textit{Impact:} Reduces transparency and explainability, making it difficult for developers and stakeholders to understand, debug, and maintain the system.  
\textit{Use Case:} Complex multi-agent systems involving multiple interacting components and evolving behaviors.  
\textit{Example:} Developers being unable to trace or explain how an agent arrived at a particular decision due to lack of documentation.

\textit{Source(s):}  \cite{liu2020using, albuquerque2022comprehending, li2023debtviz, li2023automatic, hutchinson2021towards, obrien202223, perez2021technical, nahar2022collaboration, chang2022understanding, cote2024quality, bavota2016large, liu2021exploratory, wang2023technical, li2022identifying, sutoyo2024satdaug}.

\item \textit{Cultural/People Debt:} Refers to misalignment between human users, developers, and agent behavior, including gaps in understanding, expectations, and collaboration practices.  
\textit{Impact:} Reduces trust, limits adoption, and increases the likelihood of misuse or misinterpretation of agent outputs.  
\textit{Use Case:} Human-in-the-loop systems where users interact with or supervise autonomous agents.  
\textit{Example:} Users misinterpreting agent outputs or over-trusting incorrect recommendations due to lack of clarity or understanding.

\par \textit{Source(s):} \cite{arpteg2018software, lenarduzzi2021software, perez2021technical, menshawy2024navigating, moreschini2024towards, nahar2022collaboration, mailach2023socio, annunziata2025uncovering, de2025software, akman2025people}. 

\end{enumerate}

\subsubsection{Testing and QA-Related Debts}

Debts related to insufficient validation, testing, and quality assurance of agent behaviors, interactions, and system outputs. In Agentic AI systems, these debts are particularly critical due to the dynamic, non-deterministic, and interaction-driven nature of agents. Unlike traditional software systems, where testing focuses on deterministic functions and isolated components, agentic systems require validation of complex behaviors, multi-agent coordination, and emergent outcomes across diverse scenarios.

\begin{enumerate}

\item \textit{Test Debt:} Refers to the lack of comprehensive testing strategies and validation mechanisms for agent interactions, decision-making processes, and system-level behaviors.  
\textit{Impact:} Leads to unpredictable outcomes, undetected errors, and emergent failures that may only appear during runtime or under complex conditions.  
\textit{Use Case:} Autonomous decision pipelines and multi-agent systems operating in dynamic environments.  
\textit{Example:} Agents failing under edge cases or unexpected scenarios due to insufficient testing of interaction patterns and system dynamics.

\par \textit{Source(s):}  \cite{liu2020using, tang2021empirical,   albuquerque2022comprehending, arpteg2018software, li2023debtviz, li2023automatic, lenarduzzi2021software, breck2017ml, obrien202223, perez2021technical,  shivashankar2022maintainability, bavota2016large, liu2021exploratory, wang2023technical, li2022identifying, de2025software, sutoyo2024satdaug}.

\end{enumerate}

\subsection{Summary of Key Insights}

The analysis demonstrates that AgTD extends technical debt from isolated, component-level issues to \textit{dynamic, system-level behaviors}, where interactions among agents, tools, and environments play a central role. Unlike traditional technical debt, which is typically static and localized, AgTD manifests as a \textit{distributed and interconnected phenomenon} spanning multiple agents and subsystems, thereby increasing the complexity of identification, tracking, and mitigation.

Furthermore, AgTD introduces new classes of risks driven by \textit{autonomy, coordination, and emergence}, including cascading failures, self-reinforcing feedback loops, and unpredictable system-wide behaviors. These characteristics significantly amplify the \textit{propagation of failures}, where a single defect or misalignment in one agent can rapidly affect downstream agents and compromise the entire system. At the same time, the adaptive and non-deterministic nature of agentic interactions leads to increased \textit{opacity}, making it difficult to trace root causes and understand system behavior.

As a result, technical debt management in Agentic AI systems shifts from traditional code-centric practices to \textit{behavior-centric and interaction-centric strategies}, requiring new abstractions, monitoring mechanisms, and validation approaches. In particular, the continuous evolution of agentic systems necessitates \textit{runtime governance and continuous validation}, rather than static testing and one-time verification. This also highlights the importance of integrating \textit{TRiSM principles (Trust, Risk, and Security Management)} to ensure safety, accountability, security, and regulatory compliance in autonomous multi-agent ecosystems.

In addition to its implications for software quality, trustworthiness, and governance, our synthesis indicates that AgTD also introduces an important sustainability dimension. The autonomous execution model of Agentic AI --characterized by continuous reasoning, planning, persistent memory management, tool invocation, and inter-agent coordination -- results in substantially higher computational and energy demands than traditional AI workflows. Consequently, inefficient agent architectures, redundant reasoning cycles, excessive communication overhead, persistent memory operations, and suboptimal orchestration strategies may accumulate as \emph{Sustainability Technical Debt (SusTD)}. This emerging form of debt extends beyond traditional maintenance concerns by affecting computational efficiency, operational cost, energy consumption, environmental sustainability, and the long-term viability of large-scale autonomous AI systems.

Collectively, these insights demonstrate that Agentic Technical Debt is not merely an extension of AI Technical Debt but represents a \textit{paradigm shift} in how technical debt is understood and managed. AgTD captures the transition from predictable, static systems to adaptive, distributed, and emergent ecosystems, where traditional software engineering practices alone are insufficient. These findings establish AgTD as a foundational concept for understanding and managing the trustworthiness, reliability, security, sustainability, and long-term evolution of Agentic AI systems.

\section{Implications of Agentic Technical Debt for AI TRiSM}
\label{sec:agtd_aitrism}
\begin{figure*}[t]
    \centering
    \includegraphics[width=1.0\textwidth]{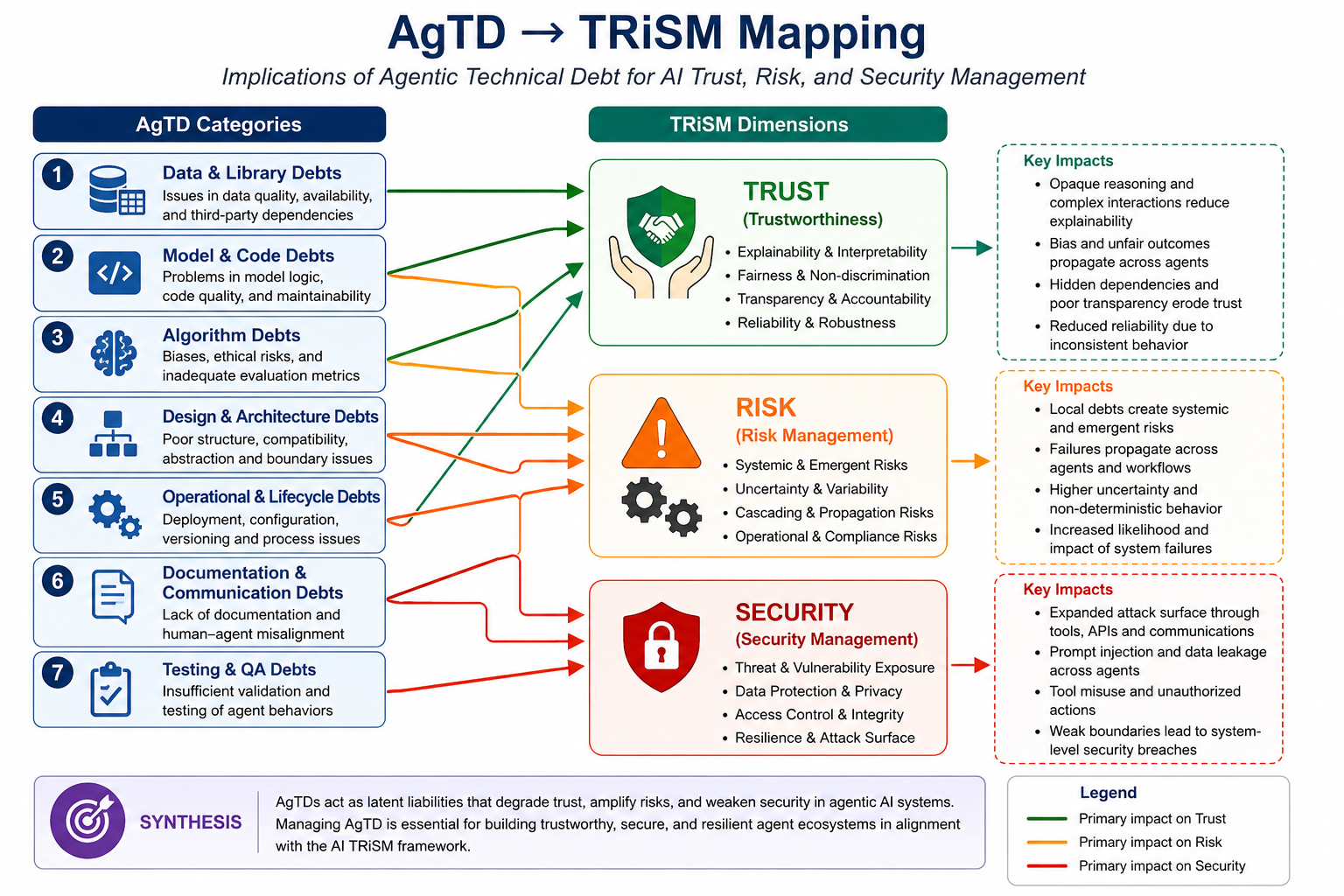}
    \caption{Mapping of AgTD categories to AI TRiSM dimensions-Trust, Risk, and Security-illustrating how different debt types degrade trustworthiness (e.g., explainability and fairness), amplify systemic and emergent risks, and expand security vulnerabilities across agentic AI systems.
}\label{fig:AgTD_AI_TRiSM_Mapping}

\end{figure*}

Building on the identified forms of Agentic Technical Debt, this section analyzes their broader implications through the lens of AI Trust, Risk, and Security Management (AI TRiSM). AI TRiSM has emerged as a governance-oriented framework that emphasizes the coordinated management of trustworthiness, risk exposure, and security resilience in AI systems~\cite{habbal2024artificial, ibm2025aitrism, nist2024aitrism, avivah2024aitrism}. While prior studies have primarily examined technical debt in relation to system quality and maintainability, AgTD introduces new challenges that extend beyond traditional engineering concerns. In agentic AI systems-characterized by autonomous, interacting agents-technical debt manifests as a systemic driver of trust degradation, risk amplification, and security vulnerabilities, as summarized in Figure~\ref{fig:AgTD_AI_TRiSM_Mapping}.

\subsection{AgTD and AI Trustworthiness}

AI TRiSM conceptualizes trust as a multi-dimensional construct encompassing reliability, transparency, explainability, fairness, and robustness~\cite{nist2024aitrism,ibm2025aitrism,raheem2025agentic}. In agentic systems, these dimensions are inherently interdependent and are certain to be significantly influenced by an accumulation of AgTD. As previously mentioned, AgTD is complex, in part, because it propagates across agent interactions, leading to emergent and system-wide degradation of trust. For instance, debts such as opaque reasoning pathways and unstructured inter-agent communication undermine explainability by obscuring decision flows across multiple agents. Similarly, coordination and memory-related debts reduce reliability, as inconsistencies in shared context or execution logic can lead to unpredictable outcomes. Biases embedded within individual agents may also propagate across the agent ecosystem, amplifying fairness concerns at the system level~\cite{bandi2025rise}. Furthermore, the dynamic and distributed nature of agentic architectures reduces transparency, as decision-making is no longer confined to a single model but distributed across multiple interacting entities~\cite{radanliev2026transparent, adabara2025trustworthy}.

In addition, the adaptive nature of agentic systems further complicates trust, as agents may evolve their behavior over time, making it difficult to ensure consistent and verifiable outcomes~\cite{raheem2025agentic}. This temporal dimension of trust degradation introduces new challenges for auditing and validation, particularly in long-running or continuously learning systems. Consequently, trust degradation in agentic AI systems is not merely a function of individual component quality but emerges from the collective behavior of interconnected agents. AgTD, if untreated, can undermine the trustworthiness of AI systems, necessitating new approaches to monitoring, auditing, and managing trust across agent ecosystems.

\subsection{AgTD as a Driver of Systemic Risk}

From a risk management perspective, AI TRiSM emphasizes the identification, assessment, and mitigation of risks arising from AI system behavior under uncertainty~\cite{ray2026review, habbal2024artificial, nist2024aitrism, avivah2024aitrism, rashid2025securing}.  AgTD significantly amplifies such risks by introducing complex dependencies, emergent behaviors, and cascading failure modes within agentic systems.
AgTD contributes to \textit{systemic risk}, where failures propagate across agents and affect the system as a whole. For example, coordination debt can lead to breakdowns in task orchestration, while memory inconsistencies may result in conflicting or erroneous decisions across agents. Similarly, feedback loops within agent interactions can reinforce suboptimal behaviors, increasing the likelihood of unintended outcomes~\cite{esen2025risks}.

A key characteristic of AgTD-driven risk is its propagation across temporal and structural dimensions. As agents continuously learn, interact, and adapt, unresolved debts accumulate and interact in non-linear ways, increasing both the likelihood and impact of system failures. This introduces a form of \textit{debt propagation risk}, where technical debt not only persists but also evolves and spreads across the agent ecosystem.
Moreover, the coupling between agents and external tools introduces additional layers of uncertainty, where failures may originate outside the system boundary yet propagate internally~\cite{kiasari2026agentic}. Therefore, AgTD is a dynamic, system-level phenomenon, requiring continuous monitoring and adaptive risk management strategies aligned with AI TRiSM principles~\cite{ray2026review, patilgoverning}.

\subsection{Security Implications of AgTD in Agentic Systems}

Security is a core pillar of AI TRiSM, focusing on protecting AI systems against adversarial threats, unauthorized access, and malicious manipulation~\cite{ibm2025aitrism, leo2026threat,  kshetri2025transforming, rashid2025securing}. Agentic AI systems inherently possess a broad attack surface due to their reliance on external tools, APIs, shared memory, and inter-agent communication channels~\cite{alva2026agentic}. When left unmanaged, AgTD can exacerbate these security challenges by embedding suboptimal design, implementation, or governance decisions into the system. For example, debts related to insecure tool usage, unvalidated inputs, or poorly governed communication protocols may increase exposure to attacks such as prompt injection, data exfiltration, and unauthorized action execution. Moreover, unresolved vulnerabilities in one agent may be more difficult to detect and remediate, increasing the risk of security issues propagating across interconnected agent workflows~\cite{gulyamov2025prompt}.


For example, debt-inducing decisions such as inadequate validation of shared memory updates, weak access-control mechanisms, or insufficient verification of external tool outputs may increase the likelihood of security incidents in agentic systems. A compromised agent may exploit these weaknesses to inject malicious information into shared memory, influencing the behavior of other agents and potentially leading to coordinated system-level failures. Similarly, insufficient validation of external tool outputs may expose decision-making processes to adversarial manipulation. These examples illustrate how unresolved AgTD can exacerbate the security challenges associated with the inherently dynamic and interconnected nature of agentic systems, making vulnerabilities more difficult to detect, trace, and remediate~\cite{evani5332681agentic}.
Furthermore, the autonomous nature of agentic systems introduces the risk of unintended actions being executed at scale, particularly when agents are granted access to external systems or resources~\cite{esen2025risks}. As a result, AgTD shifts the security paradigm from protecting isolated components to securing the entire agent ecosystem. This necessitates holistic security strategies that consider inter-agent dependencies, communication flows, and shared resources within the TRiSM framework.


\subsection{Challenges in Explainability, Governance, and System Reliability}

The emergence of AgTD introduces fundamental challenges to key TRiSM dimensions, particularly explainability, governance, and reliability. These challenges are not only amplified by the complexity of agentic systems but are also inherently intertwined, as weaknesses in one dimension often propagate and exacerbate issues in others.

\textbf{Explainability:} In agentic systems, decision-making processes often involve multiple agents operating across sequential or parallel workflows~\cite{hughes2025ai}. AgTD, if left untreated, exacerbates the opacity of these processes, making it difficult to trace how decisions are formed. Opaque reasoning chains, hidden dependencies, and dynamic interactions hinder interpretability and complicate auditing processes~\cite{janakiraman2025explainability}. In addition, the use of external tools, intermediate reasoning steps, and evolving memory states further fragments the decision trail, reducing end-to-end transparency~\cite{radanliev2026transparent}.

Moreover, explainability in agentic systems is challenged by the lack of unified representations of reasoning across agents, where each agent may employ different models, prompts, or reasoning strategies~\cite{raza2026trism}. This heterogeneity introduces inconsistencies in how decisions are generated and communicated, making it difficult to construct coherent explanations. As agents adapt over time, explanations may also become non-reproducible, further complicating validation and regulatory compliance~\cite{gruber2026foundations}. Consequently, while explainability is inherently a system-level challenge in agentic systems, unresolved AgTD can further hinder traceability, transparency, and interpretability, increasing the difficulty of understanding, validating, and governing agent behaviors across the system.

\textbf{Governance:} Effective governance requires clear accountability, well-defined control mechanisms, and enforceable policies~\cite{gangavarapu2025ai}. However, in agentic AI systems, responsibilities are distributed across multiple autonomous agents, often operating with varying levels of independence~\cite{acharya2025agentic, patilgoverning}. AgTD, if it accumulates, can complicate governance by introducing undocumented interactions, evolving system boundaries, and unclear ownership of decisions. This raises critical questions regarding accountability, compliance, and ethical responsibility, particularly in high-stakes applications.

Additionally, the dynamic nature of agentic systems challenges traditional governance models that rely on static system boundaries and predefined workflows~\cite{hughes2025ai}. Agents may dynamically invoke new tools, interact with external systems, or modify their behavior, making it difficult to enforce consistent policies. The absence of clear accountability chains---especially in multi-agent decision-making---creates ambiguity in responsibility attribution when failures occur~\cite{chabbra2025trust}. As a result, governance must evolve toward adaptive and continuous oversight mechanisms that can accommodate changing system configurations and behaviors influenced by AgTD.

\textbf{Reliability:} Reliability in agentic systems is challenged by non-deterministic execution, asynchronous interactions, and continuous adaptation~\cite{okaro2026towards}. AgTD can contributes to reliability degradation through coordination failures, inconsistent state management, and cascading errors. As agents depend on each other’s outputs, small deviations can escalate into large-scale system failures, making reliability assurance significantly more complex than in traditional systems~\cite{raheem2025agentic, xing2025looking}.

Furthermore, reliability is affected by temporal inconsistencies, where agents operate on outdated or partially synchronized information, leading to conflicting decisions. The integration of external tools and services introduces additional points of failure, often outside the direct control of the system~\cite{hughes2025ai}. AgTD can also increase the likelihood of emergent failure modes, where interactions between individually functioning agents produce unexpected system-level behaviors. Ensuring reliability in such environments requires not only robust individual components but also resilient coordination mechanisms and continuous validation of system-wide behavior.

\noindent Collectively, these challenges highlight the limitations of existing AI governance and assurance frameworks when applied to agentic systems. Traditional approaches, which focus on isolated components or static pipelines, are insufficient to address the dynamic, distributed, and evolving nature of AgTD. Addressing these challenges requires rethinking explainability, governance, and reliability as system-level properties, supported by continuous monitoring, adaptive control mechanisms, and integrated assurance frameworks tailored to agentic AI.

\subsection{Synthesis: AgTD within the AI TRiSM Framework}

Synthesizing the above analysis, AgTD can be viewed as an important contributing factor linking technical deficiencies in agentic systems to broader trust, risk, and security challenges. When left unmanaged, AgTD may reduce trust by hindering explainability and reliability, increase risk through propagation effects and emergent behaviors, and contribute to security concerns by expanding attack surfaces and weakening system boundaries.


Within the AI TRiSM framework, AgTD provides a critical lens for understanding how low-level engineering decisions translate into high-level governance and trust implications. By mapping AgTD to the core dimensions of trust, risk, and security, this study highlights the need for integrated, TRiSM-aligned approaches to technical debt management in agentic AI systems.

More importantly, AgTD emphasizes the transition from component-centric assurance to system-level governance, where trust, risk, and security must be managed holistically across interacting agents and evolving system states. This shift requires new paradigms that incorporate runtime observability, cross-agent auditing, and adaptive governance mechanisms capable of responding to dynamic system behaviors.

Ultimately, addressing AgTD is essential for enabling trustworthy, secure, and resilient agentic AI. Future research should focus on developing systematic methods for detecting, quantifying, and mitigating AgTD, as well as integrating these approaches into existing AI governance frameworks. Such efforts will be critical for bridging the gap between technical system design and high-level policy, ensuring that agentic AI systems can be deployed safely and responsibly in real-world applications.

\section{Conclusion and Future Work}
\label{sec:conclusion}

\subsection{Conclusion}

This paper addressed a critical gap in the understanding of technical debt within modern AI systems by introducing and formalizing the concept of \textit{Agentic Technical Debt (AgTD)}. While prior research on AI Technical Debt (AITD) has primarily focused on static, pipeline-oriented AI systems, the emergence of Agentic AI--characterized by autonomous reasoning, multi-agent collaboration, tool orchestration, adaptive decision-making, and persistent memory--necessitates a fundamental rethinking of how technical debt is conceptualized, analyzed, and managed.

Building upon a validated taxonomy of 31 AITDs across seven root-cause categories, this study employed a theory-informed transformation methodology to systematically reinterpret established AI technical debts within the context of Agentic AI. Through a structured transformation framework comprising direct transformation, contextual transformation, and manifestation expansion, we demonstrated how conventional component-level debts evolve into dynamic, distributed, and emergent system-level liabilities. The findings show that technical debt in Agentic AI systems is no longer confined to software artifacts, but increasingly manifests through autonomous agent behaviors, coordination mechanisms, adaptive reasoning processes, persistent memory, and interactions among agents, tools, and execution environments.

Our analysis further revealed that AgTD introduces new classes of risks, including cascading failures, coordination breakdowns, memory inconsistencies, orchestration fragility, and unsafe autonomous decision-making. Examining these risks through the lens of AI Trust, Risk, and Security Management (TRiSM) demonstrated how AgTD degrades trustworthiness, amplifies systemic risk, increases governance complexity, and expands the security attack surface. Furthermore, this work identified \emph{Sustainability Technical Debt (SusTD)} as an emerging and underexplored manifestation of AgTD, highlighting the long-term computational, energy, and environmental implications of inefficient autonomous reasoning, persistent memory management, and large-scale multi-agent orchestration.

Overall, this paper contributes to both software engineering theory and practice by:
\begin{itemize}
    \item Establishing Agentic Technical Debt (AgTD) as a foundational software engineering construct for autonomous and multi-agent AI systems.
    \item Presenting a theory-informed transformation methodology that systematically maps 31 validated AI Technical Debts to their corresponding manifestations in Agentic AI systems.
    \item Demonstrating how technical debt evolves from localized software artifacts into dynamic, behavior-centric, and interaction-driven system liabilities.
    \item Analyzing the implications of AgTD for trustworthiness, governance, security, operational resilience, and long-term software sustainability, including the emerging challenge of Sustainability Technical Debt (SusTD).
\end{itemize}

Collectively, these contributions establish AgTD as a foundational perspective for understanding and managing software quality, reliability, security, sustainability, and long-term evolution in Agentic AI systems. More broadly, this work lays the groundwork for future research on the quantification, monitoring, governance, validation, refactoring, and sustainability-aware management of technical debt in autonomous, adaptive, and multi-agent AI ecosystems.

\subsection{Future Work}

While this study establishes a conceptual foundation for Agentic Technical Debt (AgTD), several important research challenges remain open.

\renewcommand{\labelenumi}{\arabic{enumi}.}

\begin{enumerate}

\item \textbf{AgTD Quantification and Propagation Modeling:}
A key challenge is the development of formal models and metrics for quantifying AgTD and understanding how it propagates across agent interactions, shared memory, tool orchestration, and coordination workflows. Future research should investigate graph-based representations, runtime observability metrics, and probabilistic models capable of capturing debt accumulation, amplification, and cascading effects in agentic ecosystems.

\item \textbf{Runtime Detection, Monitoring, and Mitigation:}
Given the dynamic nature of agentic systems, AgTD management requires continuous runtime analysis rather than traditional static assessment techniques. Future work should focus on automated mechanisms for detecting emerging debt, monitoring interaction patterns, identifying coordination failures, and triggering adaptive mitigation or refactoring strategies during system operation.

\item \textbf{Digital Twin-Based Analysis of AgTD:}
Digital twins provide a promising approach for studying the evolution of AgTD in controlled environments. Future research should explore the use of agentic system digital twins for simulation, what-if analysis, predictive assessment of debt propagation, and evaluation of mitigation strategies before deployment in production systems.

\item \textbf{AgTD-Aware Governance and AI TRiSM Integration:}
The relationship between AgTD and Trust, Risk, and Security Management (TRiSM) warrants further investigation. Future work should develop governance frameworks, auditing mechanisms, and compliance processes that explicitly incorporate AgTD as a factor influencing explainability, accountability, reliability, and security in agentic systems.

\item \textbf{Empirical Validation in Real-World Agentic Systems:}
Finally, empirical studies are needed to validate the practical relevance of AgTD. Future research should investigate industrial case studies, longitudinal analyses, and benchmark datasets to evaluate the impact of AgTD on system quality attributes and to assess the effectiveness of proposed detection, monitoring, and mitigation approaches.

\item \textbf{Sustainability-Aware AgTD Management:}
An emerging research direction is the investigation of Sustainability Technical Debt in Agentic AI systems. Future work should develop methodologies for measuring and minimizing the energy, computational, and environmental costs associated with autonomous reasoning, persistent memory, multi-agent collaboration, and tool orchestration. This includes the development of sustainability-aware metrics, energy-efficient agent architectures, carbon footprint estimation models, workload optimization techniques~\cite{guven2026sustainability}, and Green AI engineering practices that enable long-term autonomous operation without compromising performance, reliability, or trustworthiness~\cite{cruz2025innovating, alloghani2023architecting}.

\end{enumerate}


\noindent As Agentic AI systems continue to evolve toward fully autonomous, multi-agent ecosystems, managing technical debt will become increasingly critical for ensuring trustworthiness, safety, resilience, sustainability, and long-term maintainability. This work establishes a conceptual foundation for understanding Agentic Technical Debt and outlines a research agenda encompassing its identification, quantification, monitoring, mitigation, governance, and sustainability-aware management. Addressing Agentic Technical Debt will be fundamental to developing trustworthy, efficient, and environmentally sustainable autonomous AI systems that can be safely deployed in real-world, mission-critical domains.

\bibliographystyle{abbrv}
\bibliography{ref}

@String{Computing = "Computing" }

@String{Computer = "{IEEE} Computer" }

@String{Springer = "Springer-Verlag" }

@article{tricco2018prisma,
  title={PRISMA extension for scoping reviews (PRISMA-ScR): checklist and explanation},
  author={Tricco, Andrea C and Lillie, Erin and Zarin, Wasifa and O'Brien, Kelly K and Colquhoun, Heather and Levac, Danielle and Moher, David and Peters, Micah DJ and Horsley, Tanya and Weeks, Laura and others},
  journal={Annals of internal medicine},
  volume={169},
  number={7},
  pages={467--473},
  year={2018},
  publisher={American College of Physicians},
  doi={10.7326/M18-0850}
}

@article{sculley2015hidden,
  title={Hidden technical debt in machine learning systems},
  author={Sculley, David and Holt, Gary and Golovin, Daniel and Davydov, Eugene and Phillips, Todd and Ebner, Dietmar and Chaudhary, Vinay and Young, Michael and Crespo, Jean-Francois and Dennison, Dan},
  journal={Advances in neural information processing systems},
  volume={28},
  year={2015}
}

@inproceedings{bogner2021characterizing,
  title={Characterizing technical debt and antipatterns in AI-based systems: A systematic mapping study},
  author={Bogner, Justus and Verdecchia, Roberto and Gerostathopoulos, Ilias},
  booktitle={2021 IEEE/ACM International Conference on Technical Debt (TechDebt)},
  pages={64--73},
  year={2021},
  organization={IEEE}
}

@article{recupito2024technical,
  title={Technical debt in AI-enabled systems: On the prevalence, severity, impact, and management strategies for code and architecture},
  author={Recupito, Gilberto and Pecorelli, Fabiano and Catolino, Gemma and Lenarduzzi, Valentina and Taibi, Davide and Di Nucci, Dario and Palomba, Fabio},
  journal={Journal of Systems and Software},
  volume={216},
  pages={112151},
  year={2024},
  publisher={Elsevier}
}

@article{cunningham1992wycash,
  title={The WyCash portfolio management system},
  author={Cunningham, Ward},
  journal={ACM Sigplan Oops Messenger},
  volume={4},
  number={2},
  pages={29--30},
  year={1992},
  publisher={ACM New York, NY, USA}
}

@inproceedings{alahdab2019empirical,
  title={Empirical analysis of hidden technical debt patterns in machine learning software},
  author = {Alahdab, Mohannad and Çalıklı, G{\"u}l},
  booktitle={Product-Focused Software Process Improvement: 20th International Conference, PROFES 2019, Barcelona, Spain, November 27--29, 2019, Proceedings 20},
  pages={195--202},
  year={2019},
  organization={Springer}
}

@inproceedings{chaudhary2018review,
  title={A review on hidden debts in machine learning systems},
  author={Chaudhary, Dev Kumar and Srivastava, Sandeep and Kumar, Vikas},
  booktitle={2018 Second International Conference on Green Computing and Internet of Things (ICGCIoT)},
  pages={619--624},
  year={2018},
  organization={IEEE}
}

@inproceedings{zhang2022code,
  title={Code smells for machine learning applications},
  author={Zhang, Haiyin and Cruz, Lu{\'\i}s and Van Deursen, Arie},
  booktitle={Proceedings of the 1st international conference on AI engineering: software engineering for AI},
  pages={217--228},
  year={2022}
}

@inproceedings{liu2020using,
  title={Is using deep learning frameworks free? characterizing technical debt in deep learning frameworks},
  author={Liu, Jiakun and Huang, Qiao and Xia, Xin and Shihab, Emad and Lo, David and Li, Shanping},
  booktitle={Proceedings of the ACM/IEEE 42nd International Conference on Software Engineering: Software Engineering in Society},
  pages={1--10},
  year={2020}
}

@inproceedings{10628360,
  author={Costal, Dolors and Gómez, Cristina and del Rey, Santiago and Martínez-Fernández, Silverio},
  booktitle={2024 IEEE 21st International Conference on Software Architecture Companion (ICSA-C)}, 
  title={Using Metrics for Code Smells of ML Pipelines}, 
  year={2024},
  volume={},
  number={},
  pages={289-294},
  doi={10.1109/ICSA-C63560.2024.00055}}

@inproceedings{tang2021empirical,
  title={An empirical study of refactorings and technical debt in machine learning systems},
  author={Tang, Yiming and Khatchadourian, Raffi and Bagherzadeh, Mehdi and Singh, Rhia and Stewart, Ajani and Raja, Anita},
  booktitle={2021 IEEE/ACM 43rd international conference on software engineering (ICSE)},
  pages={238--250},
  year={2021},
  organization={IEEE}
}

@article{chen2023toward,
  title={Toward understanding deep learning framework bugs},
  author={Chen, Junjie and Liang, Yihua and Shen, Qingchao and Jiang, Jiajun and Li, Shuochuan},
  journal={ACM Transactions on Software Engineering and Methodology},
  volume={32},
  number={6},
  pages={1--31},
  year={2023},
  publisher={ACM New York, NY}
}

@inproceedings{washizaki2019studying,
  title={Studying software engineering patterns for designing machine learning systems},
  author={Washizaki, Hironori and Uchida, Hiromu and Khomh, Foutse and Gu{\'e}h{\'e}neuc, Yann-Ga{\e}l},
  booktitle={2019 10th International Workshop on Empirical Software Engineering in Practice (IWESEP)},
  pages={49--495},
  year={2019},
  organization={IEEE}
}

@inproceedings{albuquerque2022comprehending,
  title={Comprehending the use of intelligent techniques to support technical debt management},
  author={Albuquerque, Danyllo and Guimaraes, Everton and Tonin, Graziela and Perkusich, Mirko and Almeida, Hyggo and Perkusich, Angelo},
  booktitle={Proceedings of the International Conference on Technical Debt},
  pages={21--30},
  year={2022}
}

@inproceedings{van2021prevalence,
  title={The prevalence of code smells in machine learning projects},
  author={Van Oort, Bart and Cruz, Lu{\'\i}s and Aniche, Maur{\'\i}cio and Van Deursen, Arie},
  booktitle={2021 IEEE/ACM 1st Workshop on AI Engineering-Software Engineering for AI (WAIN)},
  pages={1--8},
  year={2021},
  organization={IEEE}
}

@inproceedings{arpteg2018software,
  title={Software engineering challenges of deep learning},
  author={Arpteg, Anders and Brinne, Bj{\o}rn and Crnkovic-Friis, Luka and Bosch, Jan},
  booktitle={2018 44th euromicro conference on software engineering and advanced applications (SEAA)},
  pages={50--59},
  year={2018},
  organization={IEEE}
}

@inproceedings{li2023debtviz,
  title={DebtViz: A Tool for Identifying, Measuring, Visualizing, and Monitoring Self-Admitted Technical Debt},
  author={Li, Yikun and Soliman, Mohamed and Avgeriou, Paris and Van Ittersum, Maarten},
  booktitle={2023 IEEE International Conference on Software Maintenance and Evolution (ICSME)},
  pages={558--562},
  year={2023},
  organization={IEEE}
}

@article{li2023automatic,
  title={Automatic identification of self-admitted technical debt from four different sources},
  author={Li, Yikun and Soliman, Mohamed and Avgeriou, Paris},
  journal={Empirical Software Engineering},
  volume={28},
  number={3},
  pages={65},
  year={2023},
  publisher={Springer}
}

@article{polyzotis2018data,
  title={Data lifecycle challenges in production machine learning: a survey},
  author={Polyzotis, Neoklis and Roy, Sudip and Whang, Steven Euijong and Zinkevich, Martin},
  journal={ACM SIGMOD Record},
  volume={47},
  number={2},
  pages={17--28},
  year={2018},
  publisher={ACM New York, NY, USA}
}

@inproceedings{foidl2022data,
  title={Data smells: Categories, causes and consequences, and detection of suspicious data in ai-based systems},
  author={Foidl, Harald and Felderer, Michael and Ramler, Rudolf},
  booktitle={Proceedings of the 1st International Conference on AI Engineering: Software Engineering for AI},
  pages={229--239},
  year={2022}
}

@inproceedings{lenarduzzi2021software,
  title={Software quality for ai: Where we are now?},
  author={Lenarduzzi, Valentina and Lomio, Francesco and Moreschini, Sergio and Taibi, Davide and Tamburri, Damian Andrew},
  booktitle={Software Quality: Future Perspectives on Software Engineering Quality: 13th International Conference, SWQD 2021, Vienna, Austria, January 19--21, 2021, Proceedings 13},
  pages={43--53},
  year={2021},
  organization={Springer}
}

@inproceedings{foidl2019technical,
  title={Technical debt in data-intensive software systems},
  author={Foidl, Harald and Felderer, Michael and Biffl, Stefan},
  booktitle={2019 45th Euromicro conference on software engineering and advanced applications (SEAA)},
  pages={338--341},
  year={2019},
  organization={IEEE}
}

@inproceedings{breck2017ml,
  title={The ML test score: A rubric for ML production readiness and technical debt reduction},
  author={Breck, Eric and Cai, Shanqing and Nielsen, Eric and Salib, Michael and Sculley, D},
  booktitle={2017 IEEE international conference on big data (big data)},
  pages={1123--1132},
  year={2017},
  organization={IEEE}
}

@inproceedings{hutchinson2021towards,
  title={Towards accountability for machine learning datasets: Practices from software engineering and infrastructure},
  author={Hutchinson, Ben and Smart, Andrew and Hanna, Alex and Denton, Emily and Greer, Christina and Kjartansson, Oddur and Barnes, Parker and Mitchell, Margaret},
  booktitle={Proceedings of the 2021 ACM Conference on Fairness, Accountability, and Transparency},
  pages={560--575},
  year={2021}
}

@inproceedings{obrien202223,
  title={23 shades of self-admitted technical debt: An empirical study on machine learning software},
  author={OBrien, David and Biswas, Sumon and Imtiaz, Sayem and Abdalkareem, Rabe and Shihab, Emad and Rajan, Hridesh},
  booktitle={Proceedings of the 30th ACM Joint European Software Engineering Conference and Symposium on the Foundations of Software Engineering},
  pages={734--746},
  year={2022}
}

@article{perez2021technical,
  title={Technical debt payment and prevention through the lenses of software architects},
  author={P{\'e}rez, Boris and Castellanos, Camilo and Correal, Dar{\'\i}o and Rios, Nicolli and Freire, S{\'a}vio and Sp{\'\i}nola, Rodrigo and Seaman, Carolyn and Izurieta, Clemente},
  journal={Information and Software Technology},
  volume={140},
  pages={106692},
  year={2021},
  publisher={Elsevier}
}

@inproceedings{menshawy2024navigating,
  title={Navigating Challenges and Technical Debt in Large Language Models Deployment},
  author={Menshawy, Ahmed and Nawaz, Zeeshan and Fahmy, Mahmoud},
  booktitle={Proceedings of the 4th Workshop on Machine Learning and Systems},
  pages={192--199},
  year={2024}
}

@inproceedings{moreschini2024towards,
  title={Towards a Technical Debt for AI-based Recommender System},
  author={Moreschini, Sergio and Lenarduzzi, Valentina and Coba, Ludovik},
  booktitle={Proceedings of the 7th ACM/IEEE International Conference on Technical Debt},
  pages={36--39},
  year={2024}
}

@inproceedings{nahar2022collaboration,
  title={Collaboration challenges in building ml-enabled systems: Communication, documentation, engineering, and process},
  author={Nahar, Nadia and Zhou, Shurui and Lewis, Grace and K{\"a}stner, Christian},
  booktitle={Proceedings of the 44th international conference on software engineering},
  pages={413--425},
  year={2022}
}

@inproceedings{shivashankar2022maintainability,
  title={Maintainability challenges in ML: A systematic literature review},
  author={Shivashankar, Karthik and Martini, Antonio},
  booktitle={2022 48th Euromicro Conference on Software Engineering and Advanced Applications (SEAA)},
  pages={60--67},
  year={2022},
  organization={IEEE}
}

@inproceedings{chang2022understanding,
  title={Understanding implementation challenges in machine learning documentation},
  author={Chang, Jiyoo and Custis, Christine},
  booktitle={Proceedings of the 2nd ACM Conference on Equity and Access in Algorithms, Mechanisms, and Optimization},
  pages={1--8},
  year={2022}
}

@inproceedings{roselli2019managing,
  title={Managing bias in AI},
  author={Roselli, Drew and Matthews, Jeanna and Talagala, Nisha},
  booktitle={Companion proceedings of the 2019 world wide web conference},
  pages={539--544},
  year={2019}
}

@article{petrozzino2021pays,
  title={Who pays for ethical debt in AI?},
  author={Petrozzino, Catherine},
  journal={AI and Ethics},
  volume={1},
  number={3},
  pages={205--208},
  year={2021},
  publisher={Springer}
}

@article{jebnoun2022clones,
  title={Clones in deep learning code: what, where, and why?},
  author={Jebnoun, Hadhemi and Rahman, Md Saidur and Khomh, Foutse and Muse, Biruk Asmare},
  journal={Empirical Software Engineering},
  volume={27},
  number={4},
  pages={84},
  year={2022},
  publisher={Springer}
}

@article{cote2024quality,
  title={Quality issues in machine learning software systems},
  author={C{\^o}t{\'e}, Pierre-Olivier and Nikanjam, Amin and Bouchoucha, Rached and Basta, Ilan and Abidi, Mouna and Khomh, Foutse},
  journal={Empirical Software Engineering},
  volume={29},
  number={6},
  pages={1--47},
  year={2024},
  publisher={Springer}
}

@inproceedings{belani2019requirements,
  title={Requirements engineering challenges in building AI-based complex systems},
  author={Belani, Hrvoje and Vukovic, Marin and Car, {\v{Z}}eljka},
  booktitle={2019 IEEE 27th International Requirements Engineering Conference Workshops (REW)},
  pages={252--255},
  year={2019},
  organization={IEEE}
}

@inproceedings{bavota2016large,
  title={A large-scale empirical study on self-admitted technical debt},
  author={Bavota, Gabriele and Russo, Barbara},
  booktitle={Proceedings of the 13th international conference on mining software repositories},
  pages={315--326},
  year={2016}
}

@article{yan2018automating,
  title={Automating change-level self-admitted technical debt determination},
  author={Yan, Meng and Xia, Xin and Shihab, Emad and Lo, David and Yin, Jianwei and Yang, Xiaohu},
  journal={IEEE Transactions on Software Engineering},
  volume={45},
  number={12},
  pages={1211--1229},
  year={2018},
  publisher={IEEE}
}

@article{liu2021exploratory,
  title={An exploratory study on the introduction and removal of different types of technical debt in deep learning frameworks},
  author={Liu, Jiakun and Huang, Qiao and Xia, Xin and Shihab, Emad and Lo, David and Li, Shanping},
  journal={Empirical Software Engineering},
  volume={26},
  pages={1--36},
  year={2021},
  publisher={Springer}
}

@inproceedings{khritankov2021hidden,
  title={Hidden feedback loops in machine learning systems: A simulation model and preliminary results},
  author={Khritankov, Anton},
  booktitle={Software Quality: Future Perspectives on Software Engineering Quality: 13th International Conference, SWQD 2021, Vienna, Austria, January 19--21, 2021, Proceedings 13},
  pages={54--65},
  year={2021},
  organization={Springer}
}

@article{sas2023architectural,
  title={An architectural technical debt index based on machine learning and architectural smells},
  author={Sas, Darius and Avgeriou, Paris},
  journal={IEEE Transactions on Software Engineering},
  volume={49},
  number={8},
  pages={4169--4195},
  year={2023},
  publisher={IEEE}
}

@inproceedings{simon2023algorithm,
  title={Algorithm Debt: Challenges and Future Paths},
  author={Simon, Emmanuel Iko-Ojo and Vidoni, Melina and Fard, Fatemeh H},
  booktitle={2023 IEEE/ACM 2nd International Conference on AI Engineering--Software Engineering for AI (CAIN)},
  pages={90--91},
  year={2023},
  organization={IEEE}
}

@inproceedings{wang2023technical,
  title={Technical Debt Management in Industrial ML-State of Practice and Management Model Proposal},
  author={Wang, Xiaofei and Schuster, Herbert and Borrison, Reuben and Kl{\o}pper, Benjamin},
  booktitle={2023 IEEE 21st International Conference on Industrial Informatics (INDIN)},
  pages={1--9},
  year={2023},
  organization={IEEE}
}

@article{habbal2024artificial,
  title={Artificial Intelligence Trust, risk and security management (AI trism): Frameworks, applications, challenges and future research directions},
  author={Habbal, Adib and Ali, Mohamed Khalif and Abuzaraida, Mustafa Ali},
  journal={Expert Systems with Applications},
  volume={240},
  pages={122442},
  year={2024},
  publisher={Elsevier}
}

@article{avivah2024aitrism,
  title={Tackling Trust, Risk and Security in AI Models},
  author={Avivah, Litan},
  journal={https://www.gartner.com/en/articles/ai-trust-and-ai-risk},
  year={2024},
  publisher={Gartner}
}

@article{nist2024aitrism,
  title={AI Risks and Trustworthiness},
  author={NIST},
  journal={https://airc.nist.gov/airmf-resources/airmf/3-sec-characteristics/},
  year={2024},
  publisher={National Institute of Standards and Technology}
}

@inproceedings{khanvilkar2025automated,
  title={Automated Identification of Machine Learning Technical Debt Code Comments},
  author={Khanvilkar, Omkar and Mkaouer, Mohamed Wiem and AlOmar, Eman Abdullah and ElSaid, Abdelrahman and Chaaben, Amal and Touati, Mohamed},
  booktitle={2025 International Conference on Emerging Technologies and Computing (IC\_ETC)},
  pages={1--6},
  year={2025},
  organization={IEEE}
}

@inproceedings{mailach2023socio,
  title={Socio-technical anti-patterns in building ML-enabled software: insights from leaders on the forefront},
  author={Mailach, Alina and Siegmund, Norbert},
  booktitle={2023 IEEE/ACM 45th International Conference on Software Engineering (ICSE)},
  pages={690--702},
  year={2023},
  organization={IEEE}
}

@inproceedings{nikanjam2021design,
  title={Design smells in Deep Learning programs: an empirical study},
  author={Nikanjam, Amin and Khomh, Foutse},
  booktitle={2021 IEEE International conference on software maintenance and evolution (ICSME)},
  pages={332--342},
  year={2021},
  organization={IEEE}
}

@article{li2022identifying,
  title={Identifying self-admitted technical debt in issue tracking systems using machine learning},
  author={Li, Yikun and Soliman, Mohamed and Avgeriou, Paris},
  journal={Empirical Software Engineering},
  volume={27},
  number={6},
  pages={131},
  year={2022},
  publisher={Springer}
}

@inproceedings{perez2019proposed,
  title={A proposed model-driven approach to manage architectural technical debt life cycle},
  author={P{\'e}rez, Boris and Correal, Dar{\'\i}o and Astudillo, Hern{\'a}n},
  booktitle={2019 IEEE/ACM International Conference on Technical Debt (TechDebt)},
  pages={73--77},
  year={2019},
  organization={IEEE}
}

@article{annunziata2025uncovering,
  title={Uncovering community smells in machine learning-enabled systems: Causes, effects, and mitigation strategies},
  author={Annunziata, Giusy and Lambiase, Stefano and Tamburri, Damian A and Van Den Heuvel, Willem-Jan and Palomba, Fabio and Catolino, Gemma and Ferrucci, Filomena and De Lucia, Andrea},
  journal={ACM Transactions on Software Engineering and Methodology},
  volume={34},
  number={6},
  pages={1--48},
  year={2025},
  publisher={ACM New York, NY}
}

@inproceedings{recupito2024unmasking,
  title={Unmasking data secrets: An empirical investigation into data smells and their impact on data quality},
  author={Recupito, Gilberto and Rapacciuolo, Raimondo and Di Nucci, Dario and Palomba, Fabio},
  booktitle={Proceedings of the IEEE/ACM 3rd International Conference on AI Engineering-Software Engineering for AI},
  pages={53--63},
  year={2024}
}

@inproceedings{cunha2020investigating,
  title={Investigating non-usually employed features in the identification of architectural smells: A machine learning-based approach},
  author={Cunha, Warteruzannan Soyer and Armijo, Guisella Angulo and de Camargo, Valter Vieira},
  booktitle={Proceedings of the 14th Brazilian Symposium on Software Components, Architectures, and Reuse},
  pages={21--30},
  year={2020}
}

@article{de2025software,
  title={Software fairness debt: Building a research agenda for addressing bias in AI systems},
  author={de Souza Santos, Ronnie and Fronchetti, Felipe and Freire, S{\'a}vio and Spinola, Rodrigo},
  journal={ACM Transactions on Software Engineering and Methodology},
  volume={34},
  number={5},
  pages={1--21},
  year={2025},
  publisher={ACM New York, NY}
}

@inproceedings{sutoyo2024satdaug,
  title={SATDAUG-A Balanced and Augmented Dataset for Detecting Self-Admitted Technical Debt},
  author={Sutoyo, Edi and Capiluppi, Andrea},
  booktitle={Proceedings of the 21st International Conference on Mining Software Repositories},
  pages={289--293},
  year={2024}
}

@inproceedings{akgul2025aligning,
  title={Aligning Data Debt with AI-Integrated Software Project Lifecycle Processes: A Standard-Based Mapping Approach},
  author={Akg{\"u}l, Nilay Yorganc{\i}lar and Temizel, Tu{\u{g}}ba Ta{\c{s}}kaya and Top, {\"O}zden {\"O}zcan and Akman, Pelin Dayan},
  booktitle={2025 IEEE/ACM International Conference on Technical Debt (TechDebt)},
  pages={1--11},
  year={2025},
  organization={IEEE}
}

@inproceedings{ximenes2025investigating,
  title={Investigating Issues that Lead to Code Technical Debt in Machine Learning Systems},
  author={Ximenes, Rodrigo and Alves, Antonio Pedro Santos and Escovedo, Tatiana and Spinola, Rodrigo and Kalinowski, Marcos},
  booktitle={2025 IEEE/ACM 4th International Conference on AI Engineering--Software Engineering for AI (CAIN)},
  pages={173--183},
  year={2025},
  organization={IEEE}
}

@article{akman2025people,
  title={People and Management Debt in ML-Integrated Software Projects: Structuring Industry Insights},
  author={Akman, Pelin Dayan and Top, {\"O}zden {\"O}zcan and Temizel, Tugba Taskaya},
  journal={IEEE Access},
  year={2025},
  publisher={IEEE}
}

@inproceedings{shukla2022challenges,
  title={Challenges faced by industries and their potential solutions in deploying machine learning applications},
  author={Shukla, Raj Mani and Cartlidge, John},
  booktitle={2022 IEEE 12th Annual Computing and Communication Workshop and Conference (CCWC)},
  pages={0119--0124},
  year={2022},
  organization={IEEE}
}

@inproceedings{shome2022data,
  title={Data smells in public datasets},
  author={Shome, Arumoy and Cruz, Luis and Van Deursen, Arie},
  booktitle={Proceedings of the 1st International Conference on AI Engineering: Software Engineering for AI},
  pages={205--216},
  year={2022}
}

@inproceedings{moldovan2024python,
  title={The python software quality dataset},
  author={Moldovan, Vasilica-Andreea and Berciu, Liviu-Marian and Patcas, Rares-Danut},
  booktitle={2024 50th Euromicro Conference on Software Engineering and Advanced Applications (SEAA)},
  pages={395--398},
  year={2024},
  organization={IEEE}
}

@book{ernst2021technical,
  title={Technical Debt in Practice: How to Find It and Fix It},
  author={Ernst, Neil and Kazman, Rick and Delange, Julien},
  year={2021},
  publisher={MIT Press}
}

@article{abou2025agentic,
  title={Agentic AI: a comprehensive survey of architectures, applications, and future directions},
  author={Abou Ali, Mohamad and Dornaika, Fadi and Charafeddine, Jinan},
  journal={Artificial Intelligence Review},
  volume={59},
  number={1},
  pages={11},
  year={2025},
  publisher={Springer}
}

@article{RAZA202671,
title = {TRiSM for Agentic AI: A review of Trust, Risk, and Security Management in LLM-based Agentic Multi-Agent Systems},
journal = {AI Open},
volume = {7},
pages = {71-95},
year = {2026},
issn = {2666-6510},
doi = {https://doi.org/10.1016/j.aiopen.2026.02.006},
url = {https://www.sciencedirect.com/science/article/pii/S2666651026000069},
author = {Shaina Raza and Ranjan Sapkota and Manoj Karkee and Christos Emmanouilidis}
}

@inproceedings{raheem2025agentic,
  title={Agentic ai systems: Opportunities, challenges, and trustworthiness},
  author={Raheem, Tayiba and Hossain, Gahangir},
  booktitle={2025 IEEE International Conference on Electro Information Technology (eIT)},
  pages={618--624},
  year={2025},
  organization={IEEE}
}

@inproceedings{rashid2025securing,
  title={Securing Agentic AI: Threats, Risks, and Mitigation},
  author={Rashid, SM Zia Ur and Montasir, Irfanul and Haq, Ashfaqul and Ahmmed, Mohammed Tasdir and Alam, Mohammad Makchudul},
  booktitle={International Conference on Advancement In Cyber Security and Digital Forensics},
  pages={683--698},
  year={2025},
  organization={Springer}
}

@article{hosseini2025role,
  title={The role of agentic ai in shaping a smart future: A systematic review},
  author={Hosseini, Soodeh and Seilani, Hossein},
  journal={Array},
  volume={26},
  pages={100399},
  year={2025},
  publisher={Elsevier}
}

@article{islam2026rise,
  title={The rise of agentic AI: Synthesis of current knowledge and future research agenda},
  author={Islam, Md Asadul and Somu, Subbulakshmi and Aldaihani, Faraj Mazyed Faraj},
  journal={Global Business and Organizational Excellence},
  volume={45},
  number={3},
  pages={402--416},
  year={2026},
  publisher={Wiley Online Library}
}

@article{collaco2026role,
  title={The role of agentic artificial intelligence in healthcare: a scoping review},
  author={Collaco, Bernardo G and Haider, Syed Ali and Prabha, Srinivasagam and Gomez-Cabello, Cesar A and Genovese, Ariana and Wood, Nadia G and Bagaria, Sanjay P and Gopala, Narayanan and Tao, Cui and Forte, Antonio Jorge},
  journal={npj Digital Medicine},
  year={2026},
  publisher={Nature Publishing Group UK London}
}

@article{kiasari2026agentic,
  title={Agentic Artificial Intelligence for Smart Grids: A Comprehensive Review of Autonomous, Safe, and Explainable Control Frameworks},
  author={Kiasari, Mahmoud and Aly, Hamed},
  journal={Energies},
  volume={19},
  number={3},
  pages={617},
  year={2026},
  publisher={MDPI}
}

@article{acharya2025agentic,
  title={Agentic AI: Autonomous intelligence for complex goals—A comprehensive survey},
  author={Acharya, Deepak Bhaskar and Kuppan, Karthigeyan and Divya, B},
  journal={IEEe Access},
  volume={13},
  pages={18912--18936},
  year={2025},
  publisher={IEEE}
}

@article{moralles2026systematic,
  title={A Systematic Literature Review of Agentic AI: Definitions, Architectures, and Challenges},
  author={Moralles, Cassiano and Da Costa, Luis Antonio LF and Rigo, Sandro Jos{\'e} and Kunst, Rafael and De Souza, Vinicius Costa and Silva, Ederson Passos and Prado, George Lucas Ebertz and Schardosim, Taimisson De Carvalho and Roehrs, Alex},
  journal={IEEE Access},
  year={2026},
  publisher={IEEE}
}

@article{pati2025agentic,
  title={Agentic AI: a comprehensive survey of technologies, applications, and societal implications},
  author={Pati, Ashis Kumar},
  journal={IEEE Access},
  year={2025},
  publisher={IEEE}
}

@article{kostopoulos2025agentic,
  title={Agentic AI in education: State of the art and future directions},
  author={Kostopoulos, Georgios and Gkamas, Vasileios and Rigou, Maria and Kotsiantis, Sotiris},
  journal={IEEE Access},
  year={2025},
  publisher={IEEE}
}

@article{mundlamuri2025evolution,
  title={The Evolution of AI: From Classical Machine Learning to Modern Large Language Models},
  author={Mundlamuri, Rahul and Gunnam, Ganesh Reddy and Mysari, Nikhil Kumar and Pujuri, Jayakanth},
  journal={Ieee Access},
  year={2025},
  publisher={IEEE}
}

@article{gawande2025reactive,
  title={From reactive to proactive: Real-Time Human-AI collaboration in intelligent alerting systems},
  author={Gawande, Pramod Dattarao},
  journal={Journal of Computer Science and Technology Studies},
  volume={7},
  number={6},
  pages={1074--1083},
  year={2025}
}

@article{bandi2025rise,
  title={The rise of agentic ai: A review of definitions, frameworks, architectures, applications, evaluation metrics, and challenges},
  author={Bandi, Ajay and Kongari, Bhavani and Naguru, Roshini and Pasnoor, Sahitya and Vilipala, Sri Vidya},
  journal={Future Internet},
  volume={17},
  number={9},
  pages={404},
  year={2025},
  publisher={MDPI}
}

@article{chugh2025opportunities,
  title={Opportunities and Challenges of Agentic AI in Finance},
  author={Chugh, Saaniya and Deshpande, Aditya Vilas},
  journal={Journal of Emerging Technologies and Innovative Research},
  year={2025}
}

@article{kshetri2025transforming,
  title={Transforming cybersecurity with agentic AI to combat emerging cyber threats},
  author={Kshetri, Nir},
  journal={Telecommunications Policy},
  volume={49},
  number={6},
  pages={102976},
  year={2025},
  publisher={Elsevier}
}

@inproceedings{habib2025towards,
  title={Towards Explainable AI in Agentic Retrieval-Augmented Generation: A Systematic Review},
  author={Habib, Afnan and Abdulmahmod, Osamah F and Raza, Mukhlis and Gu, Yeong Hyeon and Aydo{\u{g}}an, Murat and Al-antari, Mugahed A},
  booktitle={2025 9th International Artificial Intelligence and Data Processing Symposium (IDAP)},
  pages={1--8},
  year={2025},
  organization={IEEE}
}

@article{rafe2026orchestration,
  title={Orchestration and Verification of Agentic AI Systems: A Survey of Multi-Agent Collaboration and Safety},
  author={Rafe, Isteak Ahmed and Dewan, Md Rasel and Islam, Md Reajul},
  journal={European Journal of Applied Science, Engineering and Technology},
  volume={4},
  number={2},
  pages={238--256},
  year={2026}
}

@article{leo2026threat,
  title={From threat to trust: assessing security risks of agentic AI systems: M. Leo et al.},
  author={Leo, Martin and Tan, Freedy and Miao, Tianqi and Anand, Guru},
  journal={International Journal of Information Security},
  volume={25},
  number={1},
  pages={23},
  year={2026},
  publisher={Springer}
}

@incollection{aggarwal2025traditional,
  title={Traditional ai vs modern ai},
  author={Aggarwal, Rishika and Sachan, Shachi and Verma, Rajat and Dhanda, Namrata},
  booktitle={The Confluence of Cryptography, Blockchain and Artificial Intelligence},
  pages={51--75},
  year={2025},
  publisher={CRC Press}
}

@misc{fui2023generative,
  title={Generative AI and ChatGPT: Applications, challenges, and AI-human collaboration},
  author={Fui-Hoon Nah, Fiona and Zheng, Ruilin and Cai, Jingyuan and Siau, Keng and Chen, Langtao},
  journal={Journal of information technology case and application research},
  volume={25},
  number={3},
  pages={277--304},
  year={2023},
  publisher={Taylor \& Francis}
}

@book{huang2025agentic,
  title={Agentic AI.},
  author={Huang, Ken},
  year={2025},
  publisher={Springer}
}

@article{jaboob2024artificial,
  title={Artificial intelligence: An overview},
  author={Jaboob, Ali and Durrah, Omar and Chakir, Aziza},
  journal={Engineering applications of artificial intelligence},
  pages={3--22},
  year={2024},
  publisher={Springer}
}

@article{davis1984origin,
  title={The origin of rule-based systems in AI},
  author={Davis, Randall and King, Jonathan J},
  journal={Rule-based expert systems: The MYCIN experiments of the Stanford Heuristic Programming Project},
  year={1984}
}

@article{pasrija2022machine,
  title={Machine learning and artificial intelligence: a paradigm shift in big data-driven drug design and discovery},
  author={Pasrija, Purvashi and Jha, Prakash and Upadhyaya, Pruthvi and Khan, Mohd S and Chopra, Madhu},
  journal={Current Topics in Medicinal Chemistry},
  volume={22},
  number={20},
  pages={1692--1727},
  year={2022},
  publisher={Bentham Science Publishers direct}
}

@incollection{kumar2025fundamentals,
  title={Fundamentals of Generative AI},
  author={Kumar, Krishna},
  booktitle={Generative AI for Photonic Sensing},
  pages={33--77},
  year={2025},
  publisher={Springer}
}

@incollection{parasuraman2024introduction,
  title={Introduction to generative AI and large language models (LLMs)},
  author={Parasuraman, Banu},
  booktitle={Mastering spring AI: the java developer’s guide for large language models and generative AI},
  pages={1--34},
  year={2024},
  publisher={Springer}
}

@inproceedings{kaviyaraj2024generative,
  title={Generative Artificial Intelligence: Transforming the Future},
  author={Kaviyaraj, R},
  booktitle={2024 International Conference on Emerging Technologies and Innovation for Sustainability (EmergIN)},
  pages={448--453},
  year={2024},
  organization={IEEE}
}

@article{konstantinou2024leveraging,
  title={Leveraging Generative AI Prompt Programming for Human-Robot Collaborative Assembly},
  author={Konstantinou, Christos and Antonarakos, Dimitris and Angelakis, Panagiotis and Gkournelos, Christos and Michalos, George and Makris, Sotiris},
  journal={Procedia CIRP},
  volume={128},
  pages={621--626},
  year={2024},
  publisher={Elsevier}
}

@article{garg2025designing,
  title={Designing the mind: How agentic frameworks are shaping the future of AI behavior},
  author={Garg, Venus},
  journal={Journal of Computer Science and Technology Studies},
  volume={7},
  number={5},
  pages={182--193},
  year={2025}
}

@article{widad2025evolution,
  title={The Evolution of Autonomy: From Reactive AI to Agentic AI},
  author={Widad, Jakjoud},
  journal={The Power of Agentic AI: Redefining Human Life and Decision-Making: In Industry 6.0},
  pages={1--9},
  year={2025},
  publisher={Springer}
}

@article{yang2023technical,
  title={Technical debt in the engineering of complex systems},
  author={Yang, Ye and Verma, Dinesh and Anton, Philip S},
  journal={Systems Engineering},
  volume={26},
  number={5},
  pages={590--603},
  year={2023},
  publisher={Wiley Online Library}
}

@inproceedings{brown2010managing,
  title={Managing technical debt in software-reliant systems},
  author={Brown, Nanette and Cai, Yuanfang and Guo, Yuepu and Kazman, Rick and Kim, Miryung and Kruchten, Philippe and Lim, Erin and MacCormack, Alan and Nord, Robert and Ozkaya, Ipek and others},
  booktitle={Proceedings of the FSE/SDP workshop on Future of software engineering research},
  pages={47--52},
  year={2010}
}

@article{sklavenitis2025scoping,
  title={A Scoping Review and Assessment Framework for Technical Debt in the Development and Operation of AI/ML Competition Platforms},
  author={Sklavenitis, Dionysios and Kalles, Dimitris},
  journal={Applied Sciences},
  volume={15},
  number={13},
  pages={7165},
  year={2025},
  publisher={MDPI}
}

@incollection{radanliev2026transparent,
  title={Transparent by Design: Ensuring Safety in Agentic AI Through Decision Traceability},
  author={Radanliev, Petar},
  booktitle={Ethical AI and Data Science},
  pages={1--20},
  year={2026},
  publisher={Auerbach Publications}
}

@article{adabara2025trustworthy,
  title={Trustworthy agentic AI systems: a cross-layer review of architectures, threat models, and governance strategies for real-world deployment},
  author={Adabara, Ibrahim and Sadiq, Bashir Olaniyi and Shuaibu, Aliyu Nuhu and Danjuma, Yale Ibrahim and Maninti, Venkateswarlu},
  journal={F1000Research},
  volume={14},
  number={905},
  pages={905},
  year={2025},
  publisher={F1000 Research Limited}
}

@article{ray2026review,
  title={A Review of TRiSM Frameworks in Artificial Intelligence Systems: Fundamentals, Taxonomy, Use Cases, Key Challenges and Future Directions},
  author={Ray, Partha Pratim},
  journal={Expert Systems},
  volume={43},
  number={3},
  pages={e70213},
  year={2026},
  publisher={Wiley Online Library}
}

@article{esen2025risks,
  title={The Risks of Agentic Al: The Curse of Autonomy},
  author={ESEN, Fatih Sinan},
  journal={The Age of Generative Artificial Intelligence},
  pages={156},
  year={2025},
  publisher={{\.I}zmir Akademi Derne{\u{g}}i}
}

@article{patilgoverning,
  title={Governing Agentic AI: A Strategic Framework for Autonomous Systems},
  author={Patil, Abhijeet G}
}

@article{alva2026agentic,
  title={Agentic AI systems in the age of generative models: architectures, cloud scalability, and real-world applications},
  author={Alva, Lingareddy and Pandey, Bishwajeet},
  journal={Artificial Intelligence Review},
  year={2026},
  publisher={Springer}
}

@article{gulyamov2025prompt,
  title={Prompt Injection Attacks in Large Language Models and AI Agent Systems: A Comprehensive Review of Vulnerabilities, Attack Vectors, and Defense Mechanisms},
  author={Gulyamov, Saidakhror and Gulyamov, Said and Rodionov, Andrey and Khursanov, Rustam and Mekhmonov, Kambariddin and Babaev, Djakhongir and Rakhimjonov, Akmaljon},
  year={2025},
  publisher={Preprints}
}

@article{evani5332681agentic,
  title={Agentic Ai Security: A Control Framework for Autonomous Decision-Making Systems},
  author={Evani, Pavan Kumar},
  year={2026},
  journal={Available at SSRN 5332681}
}

@article{hughes2025ai,
  title={AI agents and agentic systems: A multi-expert analysis},
  author={Hughes, Laurie and Dwivedi, Yogesh K and Malik, Tegwen and Shawosh, Mazen and Albashrawi, Mousa Ahmed and Jeon, Il and Dutot, Vincent and Appanderanda, Mandanna and Crick, Tom and De’, Rahul and others},
  journal={Journal of Computer Information Systems},
  volume={65},
  number={4},
  pages={489--517},
  year={2025},
  publisher={Taylor \& Francis}
}

@article{janakiraman2025explainability,
  title={Explainability and Interpretability in Generative AI Agents},
  author={Janakiraman, Anantharaman},
  journal={International Journal of Science, Technology and Convergence},
  volume={7},
  number={7},
  year={2025}
}

@article{raza2026trism,
  title={Trism for agentic ai: A review of trust, risk, and security management in llm-based agentic multi-agent systems},
  author={Raza, Shaina and Sapkota, Ranjan and Karkee, Manoj and Emmanouilidis, Christos},
  journal={AI Open},
  year={2026},
  publisher={Elsevier}
}

@article{gruber2026foundations,
  title={Foundations for Agentic AI Investigations from the Forensic Analysis of OpenClaw},
  author={Gruber, Jan and Hilgert, Jan-Niclas},
  journal={arXiv preprint arXiv:2604.05589},
  year={2026}
}

@incollection{gangavarapu2025ai,
  title={AI governance: preparing for the rise of Agentic AI},
  author={Gangavarapu, Rajendra},
  booktitle={Mastering AI Governance: A Guide to Building Trustworthy and Transparent AI Systems},
  pages={111--119},
  year={2025},
  publisher={Springer}
}

@article{chabbra2025trust,
  title={Trust and Accountability in Agentic AI Systems},
  author={Chabbra, Arnav},
  journal={International Journal of Computer Technology and Electronics Communication},
  volume={8},
  number={2},
  pages={10380--10387},
  year={2025}
}

@inproceedings{okaro2026towards,
  title={Towards the Reliability of Agentic Artificial Intelligence Models in Healthcare},
  author={Okaro, Ikenna Anthony and Bernardi, Mario and van der Burgt, Olaf},
  booktitle={2026 Annual Reliability and Maintainability Symposium (RAMS)},
  pages={1--6},
  year={2026},
  organization={IEEE}
}

@article{xing2025looking,
  title={Looking Forward: Challenges and Opportunities in Agentic AI Reliability},
  author={Xing, Liudong and others},
  journal={arXiv preprint arXiv:2511.11921},
  year={2025}
}

@misc{tukur2026aisafetysecuritytechnical,
      title={On AI Safety and Security Technical Debt in Engineering AI-Enabled Systems}, 
      author={Muhammad Tukur and Hayatullahi B. Adeyemo and Tao Chen and Nour Ali and Anis Zarrad and Rick Kazman and Marco Agus and Rami Bahsoon},
      year={2026},
      eprint={2607.23365},
      archivePrefix={arXiv},
      primaryClass={cs.SE},
      url={https://arxiv.org/abs/2607.23365}, 
}

@article{lee2026toward,
  title={Toward sustainable agentic ai systems: A survey of architectures and methodologies},
  author={Lee, Yubeen and Park, Eunil},
  journal={Sustainable Development},
  year={2026},
  publisher={Wiley Online Library}
}

@article{raza2025trism,
  title={Trism for agentic ai: A review of trust, risk, and security management in llm-based agentic multi-agent systems},
  author={Raza, Shaina and Sapkota, Ranjan and Karkee, Manoj and Emmanouilidis, Christos},
  journal={arXiv preprint arXiv:2506.04133},
  year={2025}
}

@article{qi2026towards,
  title={Towards trustworthy agentic AI: a comprehensive survey of safety, robustness, privacy, and system security},
  author={Qi, Jinhu and Li, Muzhi and Liu, Jiahong and Shu, Yuqin and Yu, Dianzhi and Ma, Shicheng and Cui, Wenqian and Zhao, Yiyang and Chen, Yiyi and Jiang, Ruoxi and others},
  journal={Academia AI and Applications},
  volume={2},
  number={2},
  year={2026},
  publisher={Academia. edu Journals}
}

@article{abou2026agentic,
  title={Agentic AI: a comprehensive survey of architectures, applications, and future directions},
  author={Abou Ali, Mohamad and Dornaika, Fadi and Charafeddine, Jinan},
  journal={The Artificial Intelligence Review},
  volume={59},
  number={1},
  pages={11},
  year={2026},
  publisher={Springer Nature BV}
}

@article{cruz2025innovating,
  title={Innovating for tomorrow: the convergence of software engineering and green AI},
  author={Cruz, Lu{\'\i}s and Franch, Xavier and Mart{\'\i}nez-Fern{\'a}ndez, Silverio},
  journal={ACM Transactions on Software Engineering and Methodology},
  volume={34},
  number={5},
  pages={1--13},
  year={2025},
  publisher={ACM New York, NY}
}

@incollection{alloghani2023architecting,
  title={Architecting green artificial intelligence products: Recommendations for sustainable ai software development and evaluation},
  author={Alloghani, Mohamed Ahmed},
  booktitle={Artificial Intelligence and Sustainability},
  pages={65--86},
  year={2023},
  publisher={Springer}
}

@inproceedings{guven2026sustainability,
  title={Sustainability-Aware Multi-Agent Reinforcement Learning for UAV Response and Monitoring with Joint Communication and Sensing},
  author={Guven, Islam and Parlak, Mehmet},
  booktitle={2026 IEEE Conference on Artificial Intelligence (CAI)},
  pages={1160--1165},
  year={2026},
  organization={IEEE}
}

@misc{ibm2025aitrism,
  author       = {{IBM}},
  title        = {What is AI TRiSM?},
  year         = {2025},
  url          = {https://www.ibm.com/think/topics/ai-trism},
  note         = {Accessed: 16 January 2026}
}

\end{document}